\documentclass[sigconf]{acmart}
\AtBeginDocument{%
  }

\copyrightyear{2025}
\acmYear{2025}
\setcopyright{acmlicensed}\acmConference[MM '25]{Proceedings of the 33rd ACM International Conference on Multimedia}{October 27--31, 2025}{Dublin, Ireland}
\acmBooktitle{Proceedings of the 33rd ACM International Conference on Multimedia (MM '25), October 27--31, 2025, Dublin, Ireland}
\acmDOI{10.1145/3746027.3755006}
\acmISBN{979-8-4007-2035-2/2025/10}

\usepackage{balance}
\usepackage{enumitem}
\usepackage{colortbl} 
\usepackage{multirow}
\usepackage{bbding}
\usepackage{xcolor}   
\definecolor{darkred}{rgb}{0.6, 0, 0}

\begin{document}

\title{HRSeg: High-Resolution Visual Perception and Enhancement for Reasoning Segmentation}


\author{Weihuang Lin}
\authornote{Equal contribution.}
\affiliation{%
  \department{Key Laboratory of Multimedia Trusted Perception and Efficient Computing, \\ 
  Ministry of Education of China}
  \institution{Xiamen University}
  \city{Xiamen}
  \state{Fujian}
  \country{China}
  }
\email{weihuanglin@stu.xmu.edu.cn}

\author{Yiwei Ma}
\authornotemark[1]
\affiliation{%
  \department{Key Laboratory of Multimedia Trusted Perception and Efficient Computing, \\ 
  Ministry of Education of China}
  \institution{Xiamen University}
  \city{Xiamen}
  \state{Fujian}
  \country{China}
  }
\email{yiweima@stu.xmu.edu.cn}

\author{Xiaoshuai Sun}
\authornote{Corresponding author.}
\affiliation{%
  \department{Key Laboratory of Multimedia Trusted Perception and Efficient Computing, \\ 
  Ministry of Education of China}
  \institution{Xiamen University}
  \city{Xiamen}
  \state{Fujian}
  \country{China}
  }
\email{xssun@xmu.edu.cn}

\author{Shuting He}
\affiliation{%
  \department{MoE Key Laboratory of Interdisciplinary Research of Computation and Economics}
  \institution{Shanghai University of Finance and Economics}
  \city{Shanghai}
  \country{China}
  }
\email{shuting.he@sufe.edu.cn}

\author{Jiayi Ji}
\affiliation{%
  \department{Key Laboratory of Multimedia Trusted Perception and Efficient Computing, \\ 
  Ministry of Education of China}
  \institution{Xiamen University}
  \city{Xiamen}
  \state{Fujian}
  \country{China}
  }
\email{jjyxmu@gmail.com}

\author{Liujuan Cao}
\affiliation{%
  \department{Key Laboratory of Multimedia Trusted Perception and Efficient Computing, \\ 
  Ministry of Education of China}
  \institution{Xiamen University}
  \city{Xiamen}
  \state{Fujian}
  \country{China}
  }
\email{caoliujuan@xmu.edu.cn}

\author{Rongrong Ji}
\affiliation{%
  \department{Key Laboratory of Multimedia Trusted Perception and Efficient Computing, \\ 
  Ministry of Education of China}
  \institution{Xiamen University}
  \city{Xiamen}
  \state{Fujian}
  \country{China}
  }
\email{rrji@xmu.edu.cn}

\renewcommand{\shortauthors}{Weihuang Lin et al.}

\begin{abstract}
The reasoning segmentation task involves segmenting objects within an image by interpreting implicit user instructions. Despite significant advancements made by existing approaches, they remain constrained by low perceptual resolution, as visual encoders are typically pre-trained at lower resolutions. Furthermore, simply interpolating the positional embeddings of visual encoders to enhance perceptual resolution yields only marginal performance improvements while incurring substantial computational costs. To address this, we propose HRSeg, an efficient model with high-resolution fine-grained perception. It features two key innovations: High-Resolution Perception (HRP) and High-Resolution Enhancement (HRE). The HRP module processes high-resolution images through cropping, integrating local and global features for multi-granularity quality. The HRE module enhances mask features by integrating fine-grained information from high-resolution images, refining their alignment with text features for precise segmentation. Extensive ablation studies validate the effectiveness of our modules, while comprehensive experiments on multiple benchmark datasets demonstrate HRSeg's superior performance. Code will be available at https://github.com/WeihuangLin/HRSeg.
\end{abstract}




\maketitle

\section{Introduction}

Multimodal Large Language Models (MLLMs)~\cite{wang2024qwen2,liu2024llava,liu2024oryx,bai2025qwen2,wang2023visionllm,li2023blip,chen2024internvl,li2024omg} have achieved significant advancements in vision-language understanding and reasoning. The reasoning capabilities of these models can be attributed to the exceptional performance of Large Language Models (LLMs)~\cite{openai2023gpt,touvron2023llama,jiang2024mixtral,yang2024qwen2,liu2024deepseek,guo2025deepseek}. Expanding on this, researchers~\cite{zhang2024omg,wu2024visionllm,wang2024llm,ren2024pixellm,xia2024gsva,rasheed2024glamm,liu2025seg, bai2024one, zhang2023next,yan2024visa,zheng2024villa, wei2024instructseg} are actively investigating methods to leverage reasoning capabilities of MLLMs for referring segmentation~\cite{hu2016segmentation, wang2022cris}.  This has given rise to a new task known as \textit{Reasoning Segmentation}~\cite{lai2024lisa, yang2023lisa++, liu2025seg}.

Unlike traditional referring segmentation, reasoning segmentation requires models to interpret implicit instructions to accurately identify specific targets, presenting a significantly more challenging and complex problem. Current research primarily follows two mainstream approaches.
The first category is an end-to-end approach, where reasoning and segmentation modules are simultaneously trained on a new dataset. Specifically, this method integrates LLaVA~\cite{liu2023visual} with SAM~\cite{kirillov2023segment}, enabling the model to perform both reasoning and segmentation~\cite{lai2024lisa,xia2024gsva}. This approach typically includes an additional \texttt{<SEG>} token in the LLM's vocabulary and utilizes its embedding to prompt SAM for segmentation generation. However, fine-tuning pre-trained models such as SAM requires substantial computational resources and may degrade performance due to limited training data availability.
To address these challenges, an alternative approach LLM-Seg~\cite{wang2024llm} employs a two-stage framework that separates reasoning instruction and image segmentation tasks. Initially, SAM is used to generate mask proposals, followed by selecting the correct mask using LLaVA's reasoning capability. This method relies entirely on the pre-trained SAM for generating mask proposals without fine-tuning, thereby reducing the computational resources required for training while preserving SAM's robust segmentation capability.

\begin{figure}

  \centering
  \includegraphics[width=1.\columnwidth]{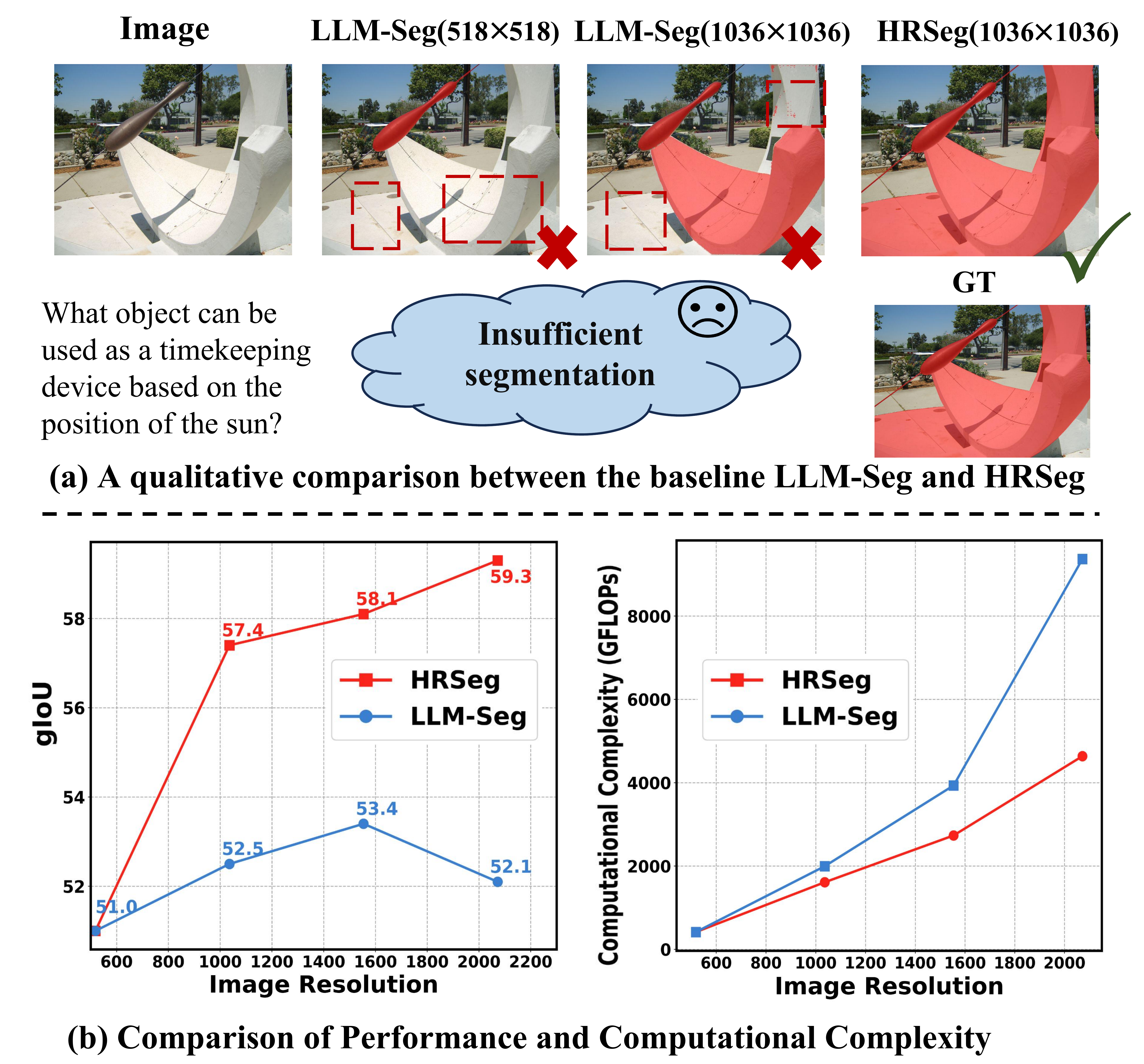}
  \vspace{-1em}
  \caption{
(a) Compared to LLM-Seg\protect\cite{wang2024llm}, which misses critical details in this complex scene, HRSeg outperforms it due to its fine-grained perception capabilities. 
(b) Comparison of performance (gIoU) and computational complexity (GFLOPs) between the two pipelines.
  }
  \vspace{-1em}
  \label{fig:intro}
\vspace{-1em}
  
\end{figure}

Despite making significant progress, the aforementioned existing works are trained on low-resolution visual encoders, such as DINOv2~\cite{zhang2022dino, oquab2023dinov2, tschannen2025siglip, bao2023all}, which are pre-trained on images at $518 \times 518$ pixels, and are therefore not inherently capable of processing high-resolution images directly.
In light of these observations, our preliminary experiments yielded two important conclusions.
\textbf{First, high-resolution image features play a crucial role in achieving accurate segmentation results.} As illustrated in the second and fourth images of Fig.~\ref{fig:intro}(a), we observe that the use of low-resolution image inputs can result in inadequate segmentation in scenarios that require fine-grained perception, an issue that our proposed HRSeg, which leverages high-resolution input, effectively addresses.
\textbf{Second, effectively processing high-resolution image inputs for reasoning segmentation remains a challenging problem.} Specifically, we introduce a naive baseline approach based on LLM-Seg for comparison, where we interpolate the pre-trained positional embeddings within the visual encoder. Although this method supports high-resolution inputs, it still leads to severe under-segmentation in complex scenarios, as evidenced by the comparison of the second and third images in Fig.~\ref{fig:intro}(a). Furthermore, as illustrated on the left side of Fig.~\ref{fig:intro}(b), this approach only improves performance within a specific resolution range; its effectiveness declines significantly as the resolution increases. More importantly, since image encoders typically rely on variants of the Vision Transformer (ViT)~\cite{dosovitskiy2020image, liu2021swin, yao2023dual, liu2021post}, their computational complexity increases quadratically with resolution, resulting in significant computational overhead, as shown on the right side of Fig.~\ref{fig:intro}(b).

To address the aforementioned challenges, we introduce an effective and innovative model for high-resolution perception and enhancement in reasoning segmentation tasks, termed High-Resolution Reasoning Segmentation (HRSeg). HRSeg integrates two novel modules to tackle visual perception difficulties in reasoning segmentation tasks.
The first module, High-Resolution Perception (HRP), conducts detailed processing of high-resolution image inputs. The high-resolution features produced are then combined with masks generated by SAM to derive mask features through interactive operations. Specifically, the input high-resolution image is cropped according to its aspect ratio and divided into sub-images to align with the pre-trained encoder's resolution. Region attention is then applied to facilitate the interaction between global image features and cropped local high-resolution features, yielding detailed high-resolution image features. Each mask generated by SAM is processed through a feature aggregation operation, referred to as mask pooling, which aggregates the image features to produce high-quality mask features.
The second module, High-Resolution Enhancement (HRE), further enriches the mask features. Since mask features are derived through mask pooling, they lack interaction with information outside the masked areas in high-resolution images. To address this, the HRE module utilizes high-resolution image features to bolster mask features. This approach enables the mask features to assimilate detailed information from high-resolution images, refining them to improve alignment with text feature \texttt{<SEG>} and enhance segmentation quality.

Based on the HRP and HRE modules, HRSeg emerges as a high-performance, computationally efficient model for the reasoning segmentation task, as shown in Fig.~\ref{fig:intro}(b), where it outperforms the naive baseline in both accuracy and efficiency. Besides, extensive experiments conducted on multiple segmentation benchmark datasets, including ReasonSeg, LLM-Seg40K, and RefCOCO/+/g, demonstrate the superior performance of HRSeg. For instance, it achieves a significant improvement of 10\% in gIoU and 15.2\% in cIoU on ReasonSeg compared to the baseline. 

In summary, the contributions of this paper are threefold:
\begin{itemize}[itemsep=0pt, topsep=0pt, parsep=0pt]
    \item We propose a novel and efficient model, HRSeg, specifically designed to address the challenges of high-resolution image perception and enhancement in the field of reasoning segmentation.
    
    \item We introduce two modules, HRP and HRE, to support the fine-grained processing pipeline of HRSeg, enabling the generation of detailed mask features.
    
    \item Extensive experiments confirm the exceptional performance of HRSeg in reasoning segmentation task.
\end{itemize}

\vspace{-10pt}
\section{Related Work}

\subsection{Multimodal Large Language Models}

With the rapid advancements in Large Language Models(LLMs)~\cite{chiang2023vicuna,touvron2023llama,zheng2023judging,team2023internlm,openai2023gpt,meta2024introducing,bi2024deepseek,yang2024qwen2}, their exceptional capabilities have garnered significant attention from researchers. Efforts have been made to extend these capabilities to the vision domain, enabling LLMs to process and understand visual information. This has spurred significant progress in the development of Multimodal Large Language Models (MLLMs)~\cite{liu2023visual,bai2023qwen,zhu2023minigpt,lu2024deepseek,liu2024oryx,li2024mini}.
MLLMs are composed of three primary components: a visual encoder, a visual projector, and an LLM. The visual encoder in MLLMs plays a critical role in extracting and interpreting visual information, with the majority of implementations adopting a Vision Transformer (ViT)-based architecture. However, its capability is constrained by low-resolution inputs, which significantly impede the MLLMs’ ability to capture fine-grained details effectively. To overcome this limitation, several researchers~\cite{ye2023ureader,lin2023sphinx,hu2024mplug,dong2024internlm,li2024monkey,li2024tokenpacker,guo2025llava,ma2024inf,huang2024mini} have employed cropping-based methods, which divide the image into multiple patches, enabling the visual encoder to effectively process high-resolution images.
The visual projector acts as a crucial intermediary between visual and textual features, facilitating the seamless integration of multimodal information. Prominent implementations of visual projectors include Q-Former~\cite{li2023blip}, Resampler~\cite{bai2023qwen}, and multilayer perceptrons (MLPs)~\cite{liu2023visual}.
%
Despite the strides made in advancing MLLMs, existing models predominantly focus on image understanding and lack strong capabilities in dense prediction tasks, such as segmentation. To bridge this gap, our research seeks to augment MLLMs by incorporating dense prediction functionalities, thus extending their application from mere image comprehension to tackling reasoning segmentation tasks.

\begin{figure*}
  \centering
  \includegraphics[width=2\columnwidth]{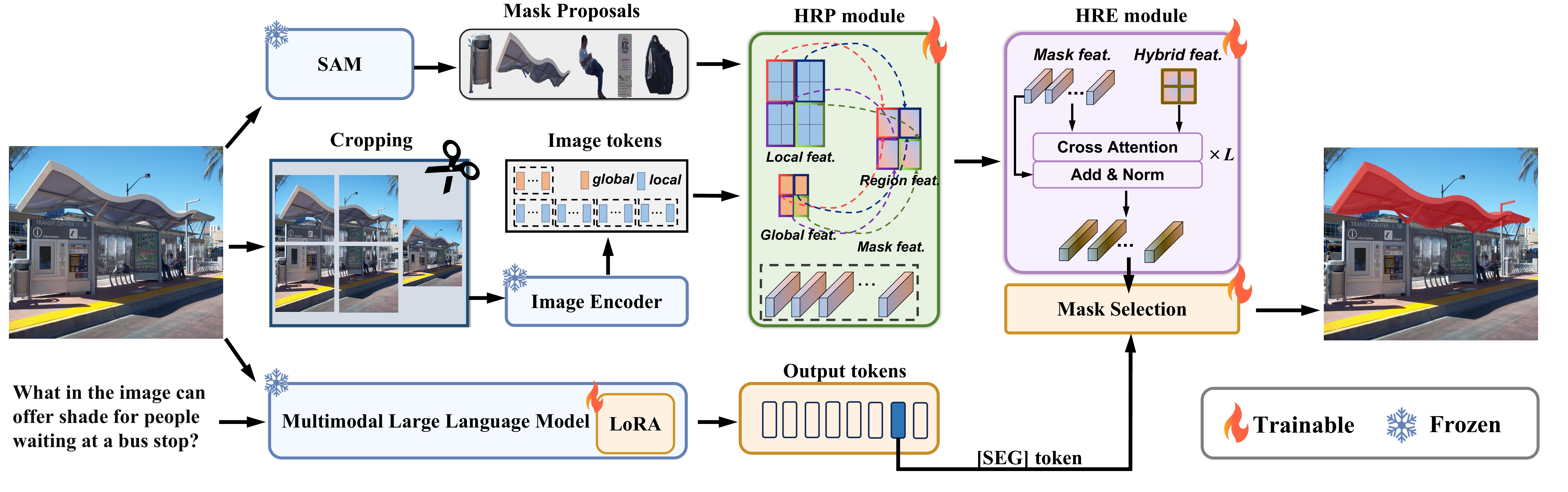}
  \vspace{-1em}
  \caption{
  The overview framework of the proposed HRSeg.
  }
  \label{fig:overview}
\vspace{-0.5em}
\end{figure*}

\subsection{Reasoning Segmentation}
The reasoning segmentation task involves generating segmentation masks guided by implicit textual instructions, making it a specialized subset of the broader referring segmentation domain. This task was initially introduced by LISA~\cite{lai2024lisa}.
Current research predominantly follows two main approaches.
The first is the end-to-end method, where reasoning and segmentation modules are jointly trained on newly curated datasets. Specifically, LISA integrates LLaVA with SAM by incorporating a \texttt{<SEG>} token into the LLM’s vocabulary, enabling SAM’s segmentation capabilities to benefit from reasoning. Building upon LISA, PixelLM~\cite{ren2024pixellm} enhances pixel-level reasoning segmentation by incorporating a segmentation codebook and a lightweight decoder, similar to SAM’s decoder, into an MLLM. In addition, LLaVASeg~\cite{yang2024empowering}, aiming to preserve the dialog capability of MLLMs, employs chain-of-thought prompting to prompt SAM, without the need to introduce a \texttt{<SEG>} token. Despite their innovations, these methods face a significant limitation: training the SAM decoder requires substantial computational resources.
As an alternative, the two-stage approach employed by LLM-Seg~\cite{wang2024llm} bridges SAM and MLLMs by selecting mask proposals without training the SAM decoder. This method successfully achieves low computational resource consumption while delivering excellent performance. Although each of these methods has its own characteristics, they do not prioritize the capture of fine-grained visual details, which is critical for dense prediction tasks. In other words, they fail to effectively perceive high-resolution images. 
In light of these insights, we propose the HRSeg model to address the need for high-resolution perception, thereby enhancing the performance of the reasoning segmentation task.

\section{Method}
In this section, we first present a comprehensive overview of the proposed HRSeg framework in Sec.~\ref{method_one}, highlighting its innovative architecture and key features. Next, we will explore the two main components of the framework: High-Resolution Perception and High-Resolution Enhancement, providing detailed explanations of their intricate functionalities in Sec.~\ref{method_two} and Sec.~\ref{method_three}, respectively. Finally, we present the training objectives of the model in Sec.~\ref{method_four}.

\subsection{Overview}
\label{method_one}
Fig.~\ref{fig:overview} illustrates the complete pipeline of the proposed HRSeg framework. To present the model architecture clearly, we divide it into two main components: \textit{Reasoning} and \textit{Segmentation}. 

The \textbf{\textit{Reasoning} component} is primarily handled by the MLLM, which comprises three key elements: a vision encoder to extract features from the image, a large language model to interpret the user’s textual instructions, and a visual projector to align the image and text features seamlessly. Following~\cite{lai2024lisa}, we adopt the embedding-as-mask paradigm by incorporating the  \texttt{<SEG>} token to the LLM’s vocabulary. Given the input image $I \in \mathbb{R}^{H \times W \times 3}$ and text instruction $T$, we process them through the MLLM to produce the output sequence. The \texttt{<SEG>} token is activated when a specific target is identified for segmentation. Subsequently, we extract the \texttt{<SEG>} token embedding and project it into the Mask Selection space:
\begin{equation}
    Y_{seg} = f_{MLLM}(I, T),
\end{equation}
\begin{equation}
    \tilde{Y}_{seg} = f_{proj}(Y_{seg}),
\end{equation}
where the function $f_{MLLM}(\cdot)$ denotes the processing carried out by MLLM, $f_{proj(\cdot)}$ denotes the MLP projector and $Y_{seg}$ corresponds to the embedding of \texttt{<SEG>} token.

In the \textbf{\textit{Segmentation} part}, the image $I$ is input to SAM, which generates mask proposals $\mathbf{M} = \{m_1, m_2, ..., m_K\} \in \mathbb{R}^{K\times D}$ for the entire image using points prompt arranged in a grid. Note that SAM focuses solely on generating pixel-accurate masks and cannot directly produce mask features. However, downstream segmentation tasks require additional feature extraction. Therefore, our goal is to integrate fine-grained image features with the mask proposals to generate high-quality mask features, enhancing the subsequent matching process. 

To leverage the image encoder trained on $I_v \in \mathbb{R} ^{H_v\times W_v\times 3}$, the image $I$ is scaled according to its aspect ratio to obtain global view of low-resolution image $I_g \in \mathbb{R}^{H_v\times W_v\times 3}$ and local view of high-resolution image $I_l \in \mathbb{R}^{(N\times H_v)\times (N\times W_v)\times 3}$, where $N$ represents the magnification. In this manner, the image encoder processes the global and local cropping image to generate corresponding image features:
\begin{equation}
    \{ \mathbf{F}_g, \mathbf{F}_l \}  = f_v(I_g, crop(I_l)),
\end{equation}
where $\mathbf{F}_g \in \mathbb{R}^{N_g\times d}$ and $\mathbf{F}_l \in \mathbb{R}^{N_l\times d}$ represent the global and local features. $N_g$ and $N_l$ denote the number of global and local tokens, respectively. The function $f_v(\cdot)$ denotes the image encoder, and $crop(\cdot)$ refers to using a window $W \in \mathbb{R}^{H_v\times W_v}$ to crop the image.

Next, the mask proposals $\mathbf{M}$ and image features $\{ \mathbf{F}_g, \mathbf{F}_l \}$ are fed into the High-Resolution Perception module, enabling the mask proposals to perceive fine-grained hybrid features:
\begin{equation}
     \{\mathbf{F}_h, \mathbf{M}_{HRP}\}  = f_{HRP}(\mathbf{M}, \mathbf{F}_g, \mathbf{F}_l),
\label{HRP}
\end{equation}
where $\mathbf{F}_h$ denotes the hybrid features, and $\mathbf{M}_{HRP} \in \mathbb{R}^{K\times d}$ represents the activated mask features. The function $f_{HRP}(\cdot)$ defines the processing performed by the HRP module, which is elaborated thoroughly in Sec.~\ref{method_two}.

Subsequently, the mask features $\mathbf{M}_{HRP}$ are enhanced through HRE module:
\begin{equation}
     \mathbf{M}_{HRE}  = f_{HRE}(\mathbf{M}_{HRP},\mathbf{F}_h),
\label{HRE}
\end{equation}
where $\mathbf{M}_{HRE}$ represents the enhanced mask features, and $f_{HRE}(\cdot)$ defines the enhancement processing, which is explained in Sec.~\ref{method_three}.
%
%
The Mask Selection module consists of a fusion module and selection heads, following the approach in ~\cite{wang2024llm}. 

Specifically, the fusion module integrates semantic information from the $\texttt{<SEG>}$ token feature $\tilde{Y}_{seg}$ with the mask features $\mathbf{M}_{HRE}$  using self-attention and cross-attention. 

The selection heads in the mask selection module consist of two separate MLPs. One maps $M_{HRE}$ into the same feature space as the text features, preparing it for the subsequent computation of the similarity score $S_{sim}$. The other is used to predict the IoP score $S_{iop}$. Firstly, calculate the similarity score  $S_{sim} \in \mathbb{R}^{1\times K}$  between the mask feature and the token feature : $S_{sim} =  \tilde{Y}_{seg} \times \mathbf{M}_{HRE}^\mathsf{T}$. Next, the mask feature $\mathbf{M}_{HRE}$ is mapped to the IoP score $S_{iop}$. This evaluates the matching degree between each mask proposal and the ground truth mask, making it particularly effective for multi-instance selection. Therefore, based on $S_{sim}$ and $S_{iop}$, we develop different strategies for mask selection. Further details are provided in the Experiments section.

\begin{equation}
     \mathbf{M}_{sel}  = f_{sel}(\mathbf{M}_{HRE},\tilde{Y}_{seg}),
\label{SEL}
\end{equation}
where the function $f_{sel}(\cdot)$ refers to the processing that combines \textit{Reasoning} and \textit{Segmentation} components.

\begin{figure}
  \centering
  \includegraphics[width=1\columnwidth]{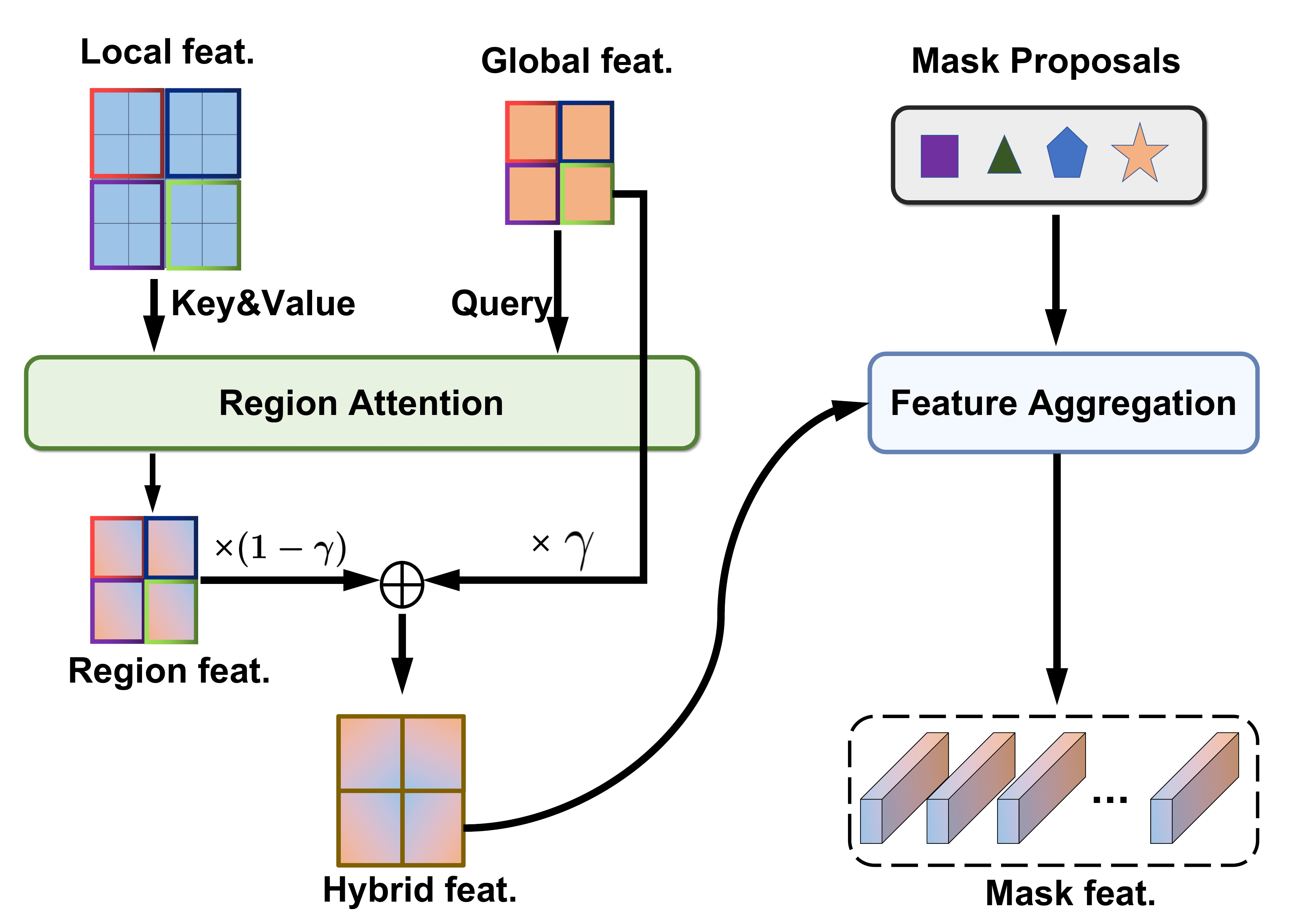}
  \vspace{-2.0em}
  \caption{
    Details of High-Resolution Perception module.
  }
  \label{fig:HRP}
\vspace{-1.5em}
\end{figure}

\begin{table*}[!h]
    \centering
    \caption{The performance of various methods on the ReasonSeg dataset. "ft" indicates whether the method has been fine-tuned using the training split of the ReasonSeg dataset.}
    \vspace{-0.4cm}
    \label{table:reason_seg}   
    \tabcolsep=0.5cm
    {
        \begin{tabular}{ l | c c | c c | c c | c c }
            \toprule
            
            \multirow{3}*{Method} & \multicolumn{2}{c|}{val} & \multicolumn{6}{c}{test} \\ 
            
            \specialrule{0em}{0pt}{1pt}
            \cline{2-9}
            \specialrule{0em}{0pt}{1pt}
            
            
            ~ & \multicolumn{2}{c|}{overall} & \multicolumn{2}{c|}{short query} & \multicolumn{2}{c|}{long query} & \multicolumn{2}{c}{overall} \\

            \specialrule{0em}{0pt}{1pt}
            \cline{2-9}
            \specialrule{0em}{0pt}{1pt}
            
            ~ & gIoU & cIoU & gIoU & cIoU & gIoU & cIoU & gIoU & cIoU \\ 
            
            \specialrule{0em}{0pt}{1pt}
            \hline
            \specialrule{0em}{0pt}{1pt}

            OVSeg~\citep{liang2023open} & 28.5 & 18.6 & 18.0 & 15.5 & 28.7 & 22.5 & 26.1 & 20.8  \\

            GRES~\citep{liu2023gres} & 22.4 & 19.9 & 17.6 & 15.0 & 22.6 & 23.8 & 21.3 & 22.0 \\    %
            
            X-Decoder~\citep{zou2023generalized} & 22.6 & 17.9 & 20.4 & 11.6 & 22.2 & 17.5 & 21.7 & 16.3 \\

            SEEM~\citep{zou2024segment} & 25.5 & 21.2 & 20.1 & 11.5 & 25.6 & 20.8 & 24.3 & 18.7 \\
            
            Grounded-SAM~\citep{liu2025grounding} & 26.0 & 14.5 & 17.8 & 10.8 & 22.4 & 18.6 & 21.3 & 16.4 \\
            
            \specialrule{0em}{0pt}{1pt}
            \hline
            \specialrule{0em}{0pt}{1pt}
            
            LISA~\citep{lai2024lisa} & 44.4 & 46.0 & 37.6 & 34.4 & 36.6 & 34.7 & 36.8 & 34.1 \\
            LLM-Seg~\citep{wang2024llm} & 47.4 & 35.4 & 39.8 & 32.9 & 41.3 & 33.8 & 40.2 & 33.4 \\
            SAM4MLLM~\citep{chen2024sam4mllm} & 46.7 & 48.1 & - & - & - & - & - & - \\
            \rowcolor{gray!10}
            HRSeg & 50.1 & 47.1 & 42.1 & 37.7 & 47.9 & 43.7 & 46.6 & 42.6 \\
            \rowcolor{gray!25}
            HRSeg-LLaVA-v1.6 & \textbf{56.7}	& \textbf{55.4}	& \textbf{47.9}	& \textbf{44.2}	& \textbf{56.6}	& \textbf{53.2}	& \textbf{54.6}	& \textbf{51.2} \\

            \specialrule{0em}{0pt}{1pt}
            \hline
            \specialrule{0em}{0pt}{1pt}

            LISA (ft) & 51.0 & 50.6 & 40.6 & 40.6 & 49.4 & 51.0 & 47.3 & 48.4 \\
            LLM-Seg (ft) & 52.3	& 47.5	& 41.1	& 40.2	& 49.8	& 49.1	& 47.9	& 46.2 \\
            \rowcolor{gray!10}
            HRSeg (ft) & 57.4 & 54.7 & 43.4 & 41.7 & 54.4 & 54.5 & 51.6 & 50.5 \\  
            \rowcolor{gray!25}
            HRSeg-LLaVA-v1.6 (ft)  & \textbf{64.9}	& \textbf{63.1}	& \textbf{49.5}	& \textbf{48.7}	& \textbf{59.2}	& \textbf{58.9}	& \textbf{57.0}	& \textbf{57.2} \\

            \bottomrule            
        \end{tabular}
    }
\vspace{-1em}
\end{table*}

\subsection{High-Resolution Perception}
\label{method_two}
Given the two types of image features, $\mathbf{F}_g$ and $\mathbf{F}_l$, the High-Resolution Perception module first enhances the global view of the image by integrating local detail features. Then, the enriched hybrid features, $\mathbf{F}_h$, are utilized to allow the mask region proposals $\mathbf{M}$ to perceive and leverage them effectively. This process is vividly illustrated in Fig.~\ref{fig:HRP}.

Firstly, we reshape $\mathbf{F}_g$ and $\mathbf{F}_l$ to get $\mathbf{F}_g \in \mathbb{R}^{n_g\times n_g\times d}$ and $\mathbf{F}_l \in \mathbb{R}^{n_l\times n_l\times d}$, where $N_g = n_g \times n_g$, $N_l = n_l \times n_l$, and $n_l = N \times n_g$. To perform region-level attention, we provide a definition to capture the region of $\mathbf{F}_l$:
\begin{equation}
    R(\mathbf{F}_l)= \mathbf{F}_l[i:i+N, j:j+N, :], 0\leq i,j\leq n_l-N,
\end{equation}
where $R(\mathbf{F}_l)\in \mathbb{R}^{(n_g\times n_g)\times (N\times N)\times d}$ represents the region features.

Next, we use $\mathbf{F}_g \in \mathbb{R}^{(n_g \times n_g) \times 1 \times d}$ as the query feature to compute the attention map between each query and the corresponding region features, enabling to identify the relevant detail region features for the query. The query feature then absorbs the high-resolution features by performing a weighted sum of the region features based on the attention maps. The region attention is formulated as follows: 
\begin{equation}
        \mathbf{F}_{r} = \text{Softmax}\Big(\frac{(\mathbf{F}_{g} \mathbf{W}^q_{g})(R(\mathbf{F}_l)\mathbf{W}^k_{l})^\mathsf{T} }{\sqrt{d}}\Big)
    (R(\mathbf{F}_l)\mathbf{W}^v_{l}),
\end{equation}
where $\mathbf{W}^q_{g}, \mathbf{W}^k_{l}, \mathbf{W}^v_{l} \in \mathbb{R}^{d \times d}$ are the trainable parameters of the projection layers. $\mathbf{F}_{r} \in \mathbb{R}^{(n_g \times n_g) \times d}$ denotes the region features.

To preserve the global perspective of the image features, we perform a pixel-wise addition of $\mathbf{F}_r$ and $\mathbf{F}_g$. This fusion yields hybrid features $\mathbf{F}_h$ that seamlessly integrate the fine-grained information from $\mathbf{F}_r$ with the holistic perspective provided by $\mathbf{F}_g$, thereby enriching the feature representation:
\begin{equation}
        \mathbf{F}_{h} = \gamma \times \mathbf{F}_g + 
                        (1 - \gamma) \times \mathbf{F}_r ,
\end{equation}
where $\gamma$ represents the weight assigned to different features.

Finally, we enable the mask proposals to perceive fine-grained hybrid features. The feature aggregation process can be formulated as follows:
\begin{equation}
    \mathbf{M}_{HRP} = \frac{\mathbf{M} \cdot \mathbf{F}^\mathsf{T}_r}{\sum \mathbf{M}},
\end{equation}
where $\mathbf{M}_{HRP} \in \mathbb{R}^{K\times d}$  denotes the activated mask feature.

\subsection{High-Resolution Enhancement}
\label{method_three}
In Sec.~\ref{method_two}, we introduced the process of enabling mask proposals to perceive high-resolution features within their internal regions. To further enhance the representation of each mask, it is essential not only to consider its own features but also to integrate the global features of the entire image. Specifically, we achieve this by leveraging the hybrid feature $\mathbf{F}_h$, a fine-grained representation that incorporates high-resolution information, to enhance the mask feature $\mathbf{M}_{HRP}$.

Specifically, to incorporate the global information of the hybrid feature into the mask features, we employ a cross-attention mechanism between them. Additionally, a residual mechanism is utilized to preserve the original mask features. The above operations are stacked for $L$ layers. This interaction can be mathematically formulated as follows:
\begin{equation}
    \mathbf{M}_{HRE} = \text{Softmax}\Big(\frac{(\mathbf{M}_{HRP} \mathbf{W}^q_{m})(\mathbf{F}_h\mathbf{W}^k_{h})^\mathsf{T} }{\sqrt{d}}\Big)
    (\mathbf{F}_h\mathbf{W}^v_{h}), 
\end{equation}
\begin{equation}
    \mathbf{M}_{HRE}= \text{Norm}(\mathbf{M}_{HRP}  + \mathbf{M}_{HRE}),
\end{equation}
where $\mathbf{W}^q_{m}, \mathbf{W}^k_{h}, \mathbf{W}^v_{h} \in \mathbb{R}^{d \times d}$ are the trainable parameters of the projection layers. The function $\text{Norm}(\cdot)$ represents layer normalization. 

Therefore, the mask feature set $\mathbf{M}_{HRE}$ contains not only fine-grained object-level information but also high-resolution features integrated throughout the image.

\subsection{Training Objectives}
\label{method_four}
The proposed model is trained using both mask selection loss $\mathcal{L}_{sel}$ and text generation loss $\mathcal{L}_{text}$.

Specifically, $\mathcal{L}_{sel}$ consists of two components: one focuses on matching the similarity and IoU distributions to select the mask that best aligns with the text, while the other component is used to supplement the mask, enabling the selection of multiple masks that better match the text under a certain threshold. This can be formally expressed as:
\begin{equation}
\mathcal{L}_{sel} = \lambda_{sim}\mathcal{L}_{sim} + \lambda_{sup}\mathcal{L}_{sup},
\end{equation}
\begin{equation}
\mathcal{L}_{sim} = \mathbf{KL}\Big( S_{sim}, \text{IoU}(\mathbf{M}_{HRE}, \mathbf{M}_{gt})
\Big),
\end{equation}
\begin{equation}
\mathcal{L}_{sup} =\text{MSE}(\text{IoP}_{pred} - \text{IoP}_{gt}),
\end{equation}
where $\mathbf{M}_{gt} \in \mathbb{R}^{1\times d}$ represents the ground-truth mask of the image. The function $\text{IoU}(\cdot)$ calculates the Intersection over Union (IoU) between $\mathbf{M}_{HRE}$ and $\mathbf{M}_{gt}$ , and $\text{KL}(\cdot)$ represents the KL divergence used to measure the two distributions. In addition, the IoP is the ratio of the area of the intersection between the ground truth and the mask proposal to the area of the proposal.

The text loss $\mathcal{L}_{text}$ is the auto-regressive cross-entropy loss used for text generation. Thus, the model’s final loss can be expressed as:
\begin{equation}
\mathcal{L} = \lambda_{text}\mathcal{L}_{text} + \lambda_{sel}\mathcal{L}_{sel}.
\end{equation}

\section{Experiments}
\subsection{Dataset and Evaluation Metric}
Following~\cite{lai2024lisa}, we train HRSeg using multiple datasets from different tasks. The first category includes semantic segmentation datasets, such as COCO-Stuff~\cite{caesar2018coco}, ADE20K~\cite{zhou2017scene}, PACO-LVIS~\cite{ramanathan2023paco}, PartImageNet~\cite{he2022partimagenet}, and PASCAL-Part~\cite{chen2014detect}. The second category includes referring segmentation datasets, namely Refclef, RefCOCO, RefCOCO+~\cite{kazemzadeh2014referitgame}, and RefCOCOg~\cite{mao2016generation}. The final category comprises reasoning segmentation datasets, including ReasonSeg~\cite{lai2024lisa} and LLM-Seg40K~\cite{wang2024llm}.

We evaluate our method using two metrics: gIoU, which computes the average IoU over all samples, and cIoU, which calculates the cumulative intersection over the cumulative union. While both metrics are used for comparison with state-of-the-art methods in our experiments, prior studies~\cite{lai2024lisa,wang2024llm} have shown that cIoU is heavily influenced by large objects. Therefore, we focus on gIoU for a more balanced evaluation.

\vspace{-1em}
\subsection{Implementation Details}
We trained HRSeg for one day using the DeepSpeed engine. The mask proposals are generated using a $32 \times 32$ grid of points to prompt SAM. The default MLLM is LLaVA-7B-v1-1, with LoRA applied for efficient fine-tuning. The image encoder used for the HRP module is DINOv2-ViT-L~\cite{oquab2023dinov2}, which is pretrained on images of size $518 \times 518$ pixels. We use AdamW~\cite{loshchilov2017decoupled} as the optimizer and apply WarmupDecayLR to adjust the learning rate. The initial learning rate is set to 0.0003. We use a batch size of 2 per GPU. For high-resolution images, we set $N =2$ by default, which corresponds to an image size of $518\times2=1036$ . The $\gamma$ in the HRP module is set to 0.8 to weight the features, and the $L$ value is set to 2 in the HRE module. For the loss function, the weights for the text loss ($\lambda_{text}$) and selection loss ($\lambda_{sel}$) are both set to 1.0.

\vspace{-10pt}
\subsection{Results on ReasonSeg}

Table~\ref{table:reason_seg} provides a detailed comparison between our proposed HRSeg and several state-of-the-art segmentation models. HRSeg is specifically designed for high-resolution reasoning segmentation tasks.

Experimental results show that HRSeg consistently outperforms existing methods across all metrics on the ReasonSeg dataset. Without fine-tuning, it surpasses LLM-Seg by 2.7\%/8.8\% (gIoU/cIoU) on the validation set and 6.4\%/9.2\% on the test set. HRSeg also handles both short and long queries better, with respective gains of 5.6\%/11.0\% and 6.6\%/9.9\%.
Fine-tuning on the ReasonSeg training split further enhances HRSeg’s performance across all evaluation metrics.

In particular, for long-query instructions that require richer visual understanding, HRSeg outperforms LLM-Seg by 6.6\%, which is 4.3\% higher than the improvement in short-query scenarios. This indicates HRSeg’s ability to capture fine-grained visual features, enabling high-quality mask generation and better alignment with complex queries.

We attribute HRSeg’s superior performance to its use of high-resolution visual information, which improves both segmentation detail and the model’s ability to handle implicit textual cues, ultimately boosting overall accuracy.

\begin{table*}[t]
    \centering
    \caption{Comparison of referring segmentation results (cIoU) between HRSeg (ours) and existing methods.}
    \label{table:refer_seg}   
    \vspace{-1em}
    \tabcolsep=0.4cm
    {
        \begin{tabular}{ l | c c c | c c c | c c }
            \toprule
            
            \multirow{2}*{Method} & \multicolumn{3}{c|}{refCOCO} & \multicolumn{3}{c|}{refCOCO+}  & \multicolumn{2}{c}{refCOCOg} \\ 
            
            \specialrule{0em}{0pt}{1pt}
            \cline{2-9}
            \specialrule{0em}{0pt}{1pt}
            
            ~ & val & testA & testB & val & testA & testB & val(U) & test(U) \\

            \specialrule{0em}{0pt}{1pt}
            \hline
            \specialrule{0em}{0pt}{1pt}
            MCN~\citep{luo2020multi} & 62.4 & 64.2 & 59.7 & 50.6 & 55.0 & 44.7 & 49.2 & 49.4 \\

            VLT~\citep{ding2021vision} & 67.5 & 70.5 & 65.2 & 56.3 & 61.0 & 50.1 & 55.0 & 57.7 \\

            CRIS~\citep{wang2022cris} & 70.5 & 73.2 & 66.1 & 62.3 & 68.1 & 53.7 & 59.9 & 60.4 \\

            LAVT~\citep{yang2022lavt} & 72.7 & 75.8 & 68.8 & 62.1 & 68.4 & 55.1 & 61.2 & 62.1 \\
            
            ReLA~\citep{liu2023gres} & 73.8 & 76.5 & 70.2 & 66.0 & 71.0 & 57.7 & 65.0 & 66.0 \\
            \hline

            LISA-LLaVA-v1~\citep{lai2024lisa}  & 74.9 & 79.1 & 72.3 & 65.1 & 70.8 & 58.1 & 67.9 & 70.6 \\
            PixelLM-LLaVA-v1~\citep{ren2024pixellm} & 73.0 & 76.5 & 68.2 & 66.3 & 71.7 & 58.3 & 69.3 & 70.5\\
            LLaVASeg-LLaVA-v1~\citep{yang2024empowering} & 76.2 & 79.1 & 72.9 & 65.7 & 71.4 & 57.7 & 69.8 & 70.1 \\
            GSVA-LLaVA-v1~\citep{xia2024gsva}	& 76.4	& 77.4	& 72.8	& 64.5	& 67.7	& 58.6	& 71.1	& 72.0 \\
            $M^2\text{SA}$-LLaVA-v1~\citep{jang2025mmr} &74.0 &76.8 &69.7 &63.1 &67.2 &56.1 &67.0 &68.3 \\
            GLaMM-LLaVA-v1.5~\citep{rasheed2024glamm} & 79.5 & 83.2 & 76.9 & 72.6 & 78.7 & 64.6 & 74.2 & 74.9 \\
            SAM4MLLM-LLaVA-v1.6~\citep{chen2024sam4mllm} & 79.6 & 82.8 & 76.1 & 73.5 & 77.8 & 65.8 & 74.5 & 75.6 \\
            SegLLM-LLaVA-v1.5~\citep{wang2024segllm} &80.2 &81.5 &75.4 &70.3 &73.0 &62.5 &72.6 &73.6 \\
            \specialrule{0em}{0pt}{1pt}
            \hline
            \specialrule{0em}{0pt}{1pt}
            
            HRSeg-LLaVA-v1 & 77.1	& 79.7	& 73.2	& 66.4	& 72.4	& 58.9	& 71.4	& 72.8 \\
            \rowcolor{gray!18}
            HRSeg-LLaVA-v1.6 & \textbf{81.2} & \textbf{83.4} & \textbf{78.1} & \textbf{73.7} & \textbf{79.1} & \textbf{67.1} & \textbf{75.6} & \textbf{76.4} \\

            \bottomrule            
        \end{tabular}
    }
    \vspace{-1em}
\end{table*}

\begin{figure*}
  \centering
  \includegraphics[width=2\columnwidth]{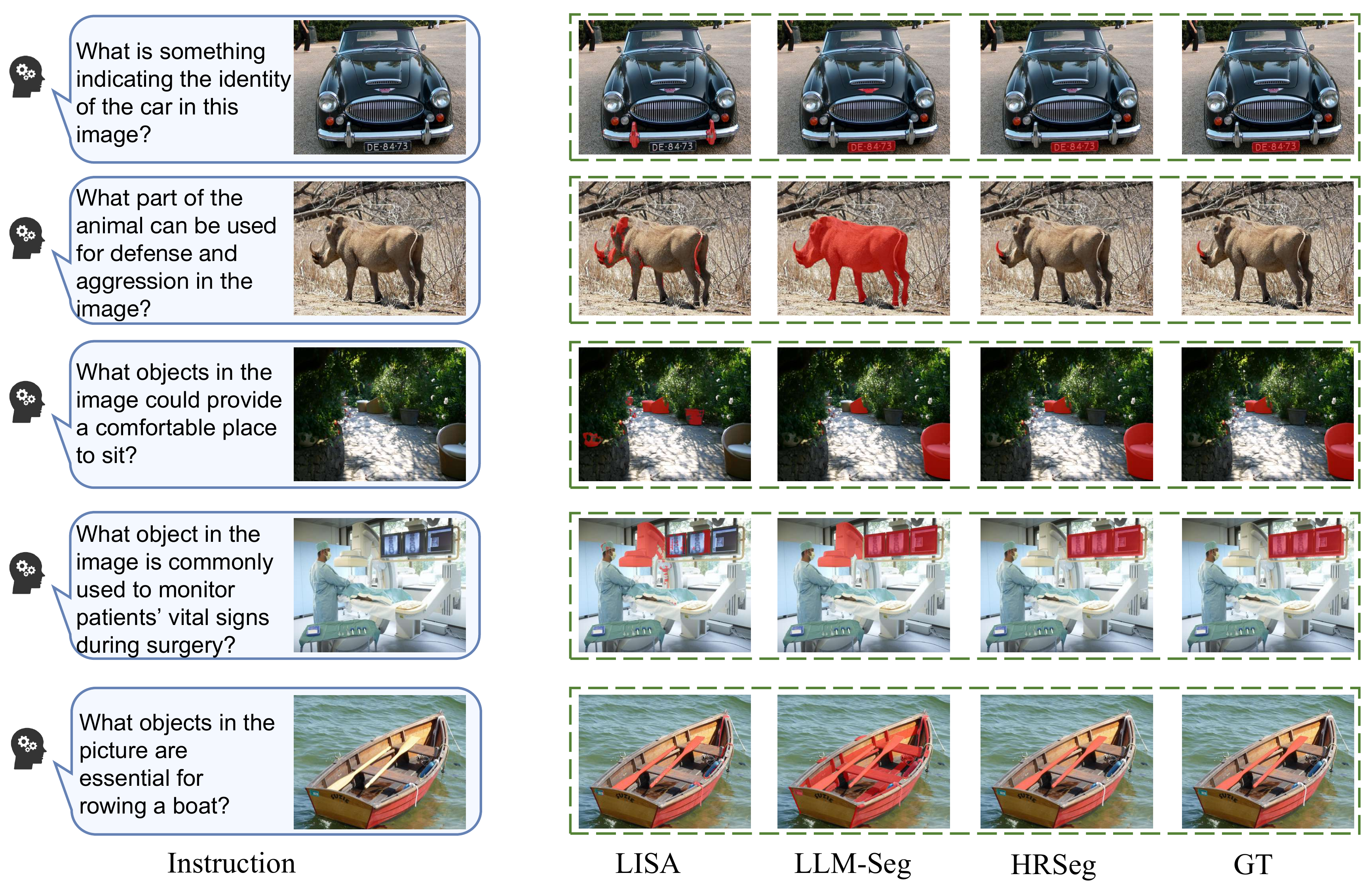}
  \vspace{-1em}
  \caption{
  Qualitative results comparison between HRSeg (Ours) and other methods. HRSeg showcases its fine-grained perception capabilities, delivering exceptional segmentation performance.
  }
  \label{fig:qualitative}
\vspace{-1.5em}
\end{figure*}

\subsection{Results on LLM-Seg40K}
The LLM-Seg40K~\cite{wang2024llm} is an additional supplement to the reasoning segmentation dataset. It is derived from LVIS~\cite{gupta2019lvis} and EgoObjects~\cite{zhu2023egoobjects}, and includes both high-quality photographic images and egocentric images.

We follow the same setting as LLM-Seg and conduct evaluation on this dataset. The results in Table~\ref{table:LLM-Seg40k} demonstrate the significant performance advantages of HRSeg over prior methods on the LLM-Seg40K dataset. Without fine-tuning, HRSeg achieves significant improvements over LLM-Seg, with gains of 8.3\% in gIoU and 5.0\% in cIoU respectively. Fine-tuning further enhances HRSeg’s performance, establishing new state-of-the-art results on LLM-Seg40K. Specifically, HRSeg achieves a remarkable 58.7 gIoU and 58.0 cIoU, outperforming fine-tuned LLM-Seg by 13.2\% in gIoU and 3.8\% in cIoU. These results highlight HRSeg’s robust ability to leverage fine-grained visual information, effectively addressing the diverse visual challenges presented by the high-quality photographic and egocentric images in the dataset.
\begin{table}[]
    
    \centering
    \caption{Performance comparison between HRSeg and prior works on the validation split of the LLM-Seg40K dataset.}
    \vspace{-1em}
    \label{table:LLM-Seg40k}
    \tabcolsep=0.9cm
    \begin{tabular}{c | c c}
        \toprule
        Method & gIoU & cIoU \\
        \midrule
    
        GRES  & 14.2 & 15.9 \\
        LISA	& 33.2	& 37.9 \\
        LLM-Seg	& 36.0	& 39.4 \\
        \rowcolor{gray!18}
        HRSeg	& \textbf{44.3} & \textbf{44.4} \\
        
        \specialrule{0em}{0pt}{1pt}
        \hline
        \specialrule{0em}{0pt}{1pt}
        
        LISA (ft)	& 37.6	& 48.5 \\
        LLM-Seg (ft)	& 45.5	& 54.2  \\
        \rowcolor{gray!18}
        HRSeg	(ft) & \textbf{58.7} & \textbf{58.0} \\

        \bottomrule
    \end{tabular}
\vspace{-1em}
\end{table}
\subsection{Results on Referring Segmentation}
Referring segmentation differs from reasoning segmentation in its use of a definite query. Table~\ref{table:refer_seg} presents the comparison results on referring segmentation datasets. Our model demonstrates competitive performance compared to recent SOTA methods, indicating its capability to effectively handle referring segmentation tasks. Furthermore, replacing MLLM with the more powerful LLaVA 1.6 leads to a notable improvement in performance.

\begin{table}[]
  \centering
  \caption{ Ablation study on the core components of HRSeg. 
  HRP and HRE denote the proposed High-Resolution Perception and High-Resolution Enhancement module.}
    \vspace{-1em}
    \setlength{\tabcolsep}{4pt}
    \begin{tabular}{c|c|cc|cc}
    \toprule
    \multirow{2}{*}{ HRP } & 
    \multirow{2}{*}{ HRE } & 
    \multicolumn{2}{c|}{  ReasonSeg val } &
    \multicolumn{2}{c}{  LLM-Seg40k val}  \\
    
    \cline { 3 -6 } 
    
    & &gIoU & cIoU & gIoU  & cIoU   \\
    \midrule
    


    & & 52.3 & 47.5 & 45.5 & 54.2 \\
    \Checkmark &  & 
    55.2\textsubscript{\textbf{\textcolor{darkred}{\scriptsize $\uparrow 5.5\%$}}} & 
    52.1\textsubscript{\textbf{\textcolor{darkred}{\scriptsize $\uparrow 9.7\%$}}} & 
    53.7\textsubscript{\textbf{\textcolor{darkred}{\scriptsize $\uparrow 18.0\%$}}} & 
    55.6\textsubscript{\textbf{\textcolor{darkred}{\scriptsize $\uparrow 2.6\%$}}} \\
    \rowcolor{gray!18}
    \Checkmark & \Checkmark & 
    \textbf{57.4}\textsubscript{\textbf{\textcolor{darkred}{\scriptsize $\uparrow 10.0\%$}}} & 
    \textbf{54.7}\textsubscript{\textbf{\textcolor{darkred}{\scriptsize $\uparrow 15.2\%$}}} & 
    \textbf{58.7}\textsubscript{\textbf{\textcolor{darkred}{\scriptsize $\uparrow 29.0\%$}}} & 
    \textbf{58.0}\textsubscript{\textbf{\textcolor{darkred}{\scriptsize $\uparrow 7.0\%$}}} \\

    \bottomrule
    \end{tabular}
    
  
\label{table:ab-component}
\vspace{-1.2em} 
\end{table}

\subsection{Ablation Study}
\noindent\textbf{Study on HRP and HRE.} 
To assess the effectiveness of our proposed HRP and HRE modules, we performed ablation studies by integrating each module into the baseline model. 
Table~\ref{table:ab-component} presents the results of the ablation study, highlighting the contributions of the HRP and HRE modules to performance improvements. The integration of the HRP module into the baseline yields significant improvements, with gIoU and cIoU metrics on the ReasonSeg validation set increasing from 52.3 to 55.2 and 47.5 to 52.1, respectively. Similarly, on the LLM-Seg40K validation set, gIoU improves from 45.5 to 53.7, and cIoU rises from 54.2 to 55.6.
Adding the HRE module alongside HRP further enhances performance, showcasing their complementary effects. The combined HRP and HRE modules achieve the highest metrics, with a gIoU of 57.4 and cIoU of 54.7 on the ReasonSeg validation set. These results underscore the HRP module’s strength in high-resolution feature perception and the HRE module’s role in refining these features, collectively driving state-of-the-art segmentation performance.

\noindent\textbf{Study on the weight $\gamma$.} 
To assess the effect of weighting global and region features, we analyze different values of $\gamma$ in the HRP module. As shown in Table~\ref{table:ab-gamma}, $\gamma$ significantly impacts the model’s ability to integrate these feature types. A low $\gamma$ weakens global context modeling, while a high value suppresses crucial region-specific details. The optimal setting, $\gamma = 0.8$, strikes a balance, effectively combining global and local cues to improve reasoning segmentation performance.

\begin{table}[]
  \centering
  \caption{ Ablation study on the weight $\gamma$ in the HRP module. }
    \vspace{-1em}
    \setlength{\tabcolsep}{1.0cm}
    \begin{tabular}{c|cc}
    \toprule
    \multirow{2}{*}{ $\gamma$ } & 
    \multicolumn{2}{c}{ ReasonSeg val}  \\
    \cline { 2 -3 } 
    & gIoU & cIoU    \\
    \midrule
     0.5 &  56.3 & 53.9  \\
     0.6 &  56.5 & 54.1  \\
     0.7 &  57.1 & 54.3  \\
     \rowcolor{gray!18}
     0.8 &  \textbf{57.4} & \textbf{54.7}  \\ 
     0.9 &  56.9 & 54.4  \\
    \bottomrule
    \end{tabular}
    \vspace{-1em}
\label{table:ab-gamma}
\end{table}

\begin{table}[]
  \centering
  \caption{ Ablation study on the size of high-resolution images. }
    \vspace{-1em}
    \tabcolsep=0.5cm
    \begin{tabular}{c|cc|cc}
    \toprule
    \multirow{2}{*}{ N } & 
    \multicolumn{2}{c|}{  ReasonSeg val } &
    \multicolumn{2}{c}{  LLM-Seg40k val}  \\
    \cline { 2 -5 } 
    & gIoU & cIoU & gIoU  & cIoU   \\
    \midrule
     - &  52.3 & 47.5 & 45.5 & 54.2 \\
     2 &  57.4 & 54.7 & 58.7 & 58.0 \\
     3 &  58.1 & 54.5 & 61.2 & 57.3 \\
     4 &  59.3 & 55.3 & 62.1 & 58.2 \\ 
    \bottomrule
    \end{tabular}
\vspace{-2em} 
\label{table:ab-resolution}
\end{table}

\noindent\textbf{Study on the image resolution.}
To further investigate the impact of increased image resolution on the performance of our model, we conducted an ablation study on image resolution.
As shown in Table~\ref{table:ab-resolution}, $N$ denotes the magnification factor relative to the pre-trained encoder’s input. Results show that higher resolutions improve segmentation by capturing finer details, benefiting both global and local reasoning. However, the performance gains taper off as resolution increases, while computational cost rises sharply. Therefore, we select $N=2$ as the optimal setting, offering a favorable trade-off between accuracy and efficiency.

\noindent\textbf{Study on the mask selection strategy.}
To investigate the impact of mask selection strategies on final segmentation performance, we experimented with different strategies. 
As shown in Table~\ref{table:ab-strategy}, we first evaluated single-mask selection using either the highest $S_{sim}$ or $S_{iop}$ score, but this approach performed significantly worse than multi-mask strategies. Among the five options, the most effective strategy considers masks with high semantic similarity to text features while maximizing those with high $S_{iop}$ scores. This approach achieved 56.9 gIoU and 54.4 cIoU on the ReasonSeg validation set. By combining multiple over-segmented SAM proposals, this strategy enables HRSeg to reach optimal performance. Furthermore, we conducted experiments on the threshold of $S_{iop}$. As shown in Table~\ref{table:ab-threshold-iop}, the initial threshold of 0.5 did not yield the best performance. Instead, the optimal performance was achieved when the threshold was set to 0.7. This suggests that the initial setting included some low-confidence masks, which negatively impacted segmentation performance.

\begin{table}[]
  \centering
  \caption{ Ablation study on the mask selection strategy In Multi-mask selection strategy, the threshold is 0.5 and K is 10.}
    \vspace{-1em}
    \setlength{\tabcolsep}{0.4cm}
    \begin{tabular}{c|cc}
    \toprule
    \multirow{2}{*}{ Mask selection strategy } & 
    \multicolumn{2}{c}{ ReasonSeg val}  \\
    \cline{2-3} 
    & gIoU & cIoU    \\
    \midrule
    \rowcolor{gray!18} \multicolumn{3}{c}{Single-mask selection strategy} \\
    \midrule
    
    Top1 $S_{sim}$ & 52.4 & 51.9 \\
    Top1 $S_{IoP}$ & 51.9 & 50.8 \\
    \midrule
    \rowcolor{gray!18} \multicolumn{3}{c}{Multi-mask selection strategy} \\
    \midrule
    
    $S_{IoP}$ from threshold & 55.1 & 52.6  \\
    Top K $S_{sim}$ & 54.2 & 53.1  \\
    \rowcolor{gray!18}
    Top K $S_{sim} \cap S_{IoP}$ from threshold   &  \textbf{56.9} & \textbf{54.4}  \\
    \bottomrule
    \end{tabular}
\vspace{-1em} 
\label{table:ab-strategy}
\end{table}

\begin{table}[]
  \centering
  \caption{ Ablation study on the threshold of $S_{IoP}$ }
    \vspace{-1em}
    \setlength{\tabcolsep}{0.9cm}
    \begin{tabular}{c|cc}
    \toprule
    \multirow{2}{*}{ threshold } & 
    \multicolumn{2}{c}{ ReasonSeg val}  \\
    \cline { 2 -3 } 
    & gIoU & cIoU    \\
    \midrule
     0.5 &  56.9 & 54.4  \\
     0.6 &  57.1 & 54.5  \\
     \rowcolor{gray!18}
     \textbf{0.7} &  \textbf{57.4} & \textbf{54.7}  \\
     0.8 &  56.7 & 54.2  \\ 
     0.9 &  55.9 & 53.8  \\
    \bottomrule
    \end{tabular}
\label{table:ab-threshold-iop}
\vspace{-2em} 
\end{table}

\vspace{-1em}
\subsection{Qualitative Results}
Fig.~\ref{fig:qualitative} shows qualitative results on the ReasonSeg dataset, comparing HRSeg with other SOTA methods. HRSeg consistently produces more accurate and detailed segmentations. In the first three rows, it performs especially well on small objects and visually demanding scenarios. This is attributed to HRSeg’s key components that enhance fine-grained perception, allowing it to handle implicit instructions and achieve precise segmentation. In contrast, other methods often fail in complex cases, yielding inaccurate or coarse results.

\vspace{-1em}
\section{Conclusion}
In this paper, we introduce HRSeg, an advanced framework specifically designed to capture fine-grained visual details and address the challenges of high-resolution reasoning segmentation tasks. Through comprehensive experimental evaluations, we demonstrate that the proposed HRP and HRE modules enhance segmentation performance across various benchmark datasets. These results underscore the effectiveness and adaptability of our framework, establishing it as a robust solution for tackling complex real-world segmentation scenarios. We hope HRSeg will serve as a strong foundation for future innovations in reasoning segmentation task.

\begin{acks}
This work was supported by National Key R\&D Program of China (No.2023YFB4502804), the National Science Fund for Distinguished Young Scholars (No.62025603), the National Natural Science Foundation of China (No. U22B2051, No. U21B2037, No. 62302411, No. 624B2118), the Natural Science Foundation of Fujian Province of China (No.2021J06003), and China Postdoctoral Science Foundation (No. 2023M732948).
\end{acks}

\bibliographystyle{ACM-Reference-Format}
\balance
\bibliography{HRSeg}


\begin{thebibliography}{80}


\ifx \showCODEN    \undefined \def \showCODEN     #1{\unskip}     \fi
\ifx \showISBNx    \undefined \def \showISBNx     #1{\unskip}     \fi
\ifx \showISBNxiii \undefined \def \showISBNxiii  #1{\unskip}     \fi
\ifx \showISSN     \undefined \def \showISSN      #1{\unskip}     \fi
\ifx \showLCCN     \undefined \def \showLCCN      #1{\unskip}     \fi
\ifx \shownote     \undefined \def \shownote      #1{#1}          \fi
\ifx \showarticletitle \undefined \def \showarticletitle #1{#1}   \fi
\ifx \showURL      \undefined \def \showURL       {\relax}        \fi
\providecommand\bibfield[2]{#2}
\providecommand\bibinfo[2]{#2}
\providecommand\natexlab[1]{#1}
\providecommand\showeprint[2][]{arXiv:#2}

\bibitem[Bai et~al\mbox{.}(2023)]%
        {bai2023qwen}
\bibfield{author}{\bibinfo{person}{Jinze Bai}, \bibinfo{person}{Shuai Bai}, \bibinfo{person}{Shusheng Yang}, \bibinfo{person}{Shijie Wang}, \bibinfo{person}{Sinan Tan}, \bibinfo{person}{Peng Wang}, \bibinfo{person}{Junyang Lin}, \bibinfo{person}{Chang Zhou}, {and} \bibinfo{person}{Jingren Zhou}.} \bibinfo{year}{2023}\natexlab{}.
\newblock \showarticletitle{Qwen-vl: A frontier large vision-language model with versatile abilities}.
\newblock \bibinfo{journal}{\emph{arXiv preprint arXiv:2308.12966}} (\bibinfo{year}{2023}).
\newblock


\bibitem[Bai et~al\mbox{.}(2025)]%
        {bai2025qwen2}
\bibfield{author}{\bibinfo{person}{Shuai Bai}, \bibinfo{person}{Keqin Chen}, \bibinfo{person}{Xuejing Liu}, \bibinfo{person}{Jialin Wang}, \bibinfo{person}{Wenbin Ge}, \bibinfo{person}{Sibo Song}, \bibinfo{person}{Kai Dang}, \bibinfo{person}{Peng Wang}, \bibinfo{person}{Shijie Wang}, \bibinfo{person}{Jun Tang}, {et~al\mbox{.}}} \bibinfo{year}{2025}\natexlab{}.
\newblock \showarticletitle{Qwen2. 5-vl technical report}.
\newblock \bibinfo{journal}{\emph{arXiv preprint arXiv:2502.13923}} (\bibinfo{year}{2025}).
\newblock


\bibitem[Bai et~al\mbox{.}(2024)]%
        {bai2024one}
\bibfield{author}{\bibinfo{person}{Zechen Bai}, \bibinfo{person}{Tong He}, \bibinfo{person}{Haiyang Mei}, \bibinfo{person}{Pichao Wang}, \bibinfo{person}{Ziteng Gao}, \bibinfo{person}{Joya Chen}, \bibinfo{person}{Zheng Zhang}, {and} \bibinfo{person}{Mike~Zheng Shou}.} \bibinfo{year}{2024}\natexlab{}.
\newblock \showarticletitle{One token to seg them all: Language instructed reasoning segmentation in videos}.
\newblock \bibinfo{journal}{\emph{Advances in Neural Information Processing Systems}}  \bibinfo{volume}{37} (\bibinfo{year}{2024}), \bibinfo{pages}{6833--6859}.
\newblock


\bibitem[Bao et~al\mbox{.}(2023)]%
        {bao2023all}
\bibfield{author}{\bibinfo{person}{Fan Bao}, \bibinfo{person}{Shen Nie}, \bibinfo{person}{Kaiwen Xue}, \bibinfo{person}{Yue Cao}, \bibinfo{person}{Chongxuan Li}, \bibinfo{person}{Hang Su}, {and} \bibinfo{person}{Jun Zhu}.} \bibinfo{year}{2023}\natexlab{}.
\newblock \showarticletitle{All are worth words: A vit backbone for diffusion models}. In \bibinfo{booktitle}{\emph{Proceedings of the IEEE/CVF conference on computer vision and pattern recognition}}. \bibinfo{pages}{22669--22679}.
\newblock


\bibitem[Bi et~al\mbox{.}(2024)]%
        {bi2024deepseek}
\bibfield{author}{\bibinfo{person}{Xiao Bi}, \bibinfo{person}{Deli Chen}, \bibinfo{person}{Guanting Chen}, \bibinfo{person}{Shanhuang Chen}, \bibinfo{person}{Damai Dai}, \bibinfo{person}{Chengqi Deng}, \bibinfo{person}{Honghui Ding}, \bibinfo{person}{Kai Dong}, \bibinfo{person}{Qiushi Du}, \bibinfo{person}{Zhe Fu}, {et~al\mbox{.}}} \bibinfo{year}{2024}\natexlab{}.
\newblock \showarticletitle{Deepseek llm: Scaling open-source language models with longtermism}.
\newblock \bibinfo{journal}{\emph{arXiv preprint arXiv:2401.02954}} (\bibinfo{year}{2024}).
\newblock


\bibitem[Caesar et~al\mbox{.}(2018)]%
        {caesar2018coco}
\bibfield{author}{\bibinfo{person}{Holger Caesar}, \bibinfo{person}{Jasper Uijlings}, {and} \bibinfo{person}{Vittorio Ferrari}.} \bibinfo{year}{2018}\natexlab{}.
\newblock \showarticletitle{Coco-stuff: Thing and stuff classes in context}. In \bibinfo{booktitle}{\emph{Proceedings of the IEEE conference on computer vision and pattern recognition}}. \bibinfo{pages}{1209--1218}.
\newblock


\bibitem[Chen et~al\mbox{.}(2014)]%
        {chen2014detect}
\bibfield{author}{\bibinfo{person}{Xianjie Chen}, \bibinfo{person}{Roozbeh Mottaghi}, \bibinfo{person}{Xiaobai Liu}, \bibinfo{person}{Sanja Fidler}, \bibinfo{person}{Raquel Urtasun}, {and} \bibinfo{person}{Alan Yuille}.} \bibinfo{year}{2014}\natexlab{}.
\newblock \showarticletitle{Detect what you can: Detecting and representing objects using holistic models and body parts}. In \bibinfo{booktitle}{\emph{Proceedings of the IEEE conference on computer vision and pattern recognition}}. \bibinfo{pages}{1971--1978}.
\newblock


\bibitem[Chen et~al\mbox{.}(2024a)]%
        {chen2024sam4mllm}
\bibfield{author}{\bibinfo{person}{Yi-Chia Chen}, \bibinfo{person}{Wei-Hua Li}, \bibinfo{person}{Cheng Sun}, \bibinfo{person}{Yu-Chiang~Frank Wang}, {and} \bibinfo{person}{Chu-Song Chen}.} \bibinfo{year}{2024}\natexlab{a}.
\newblock \showarticletitle{SAM4MLLM: Enhance Multi-Modal Large Language Model for Referring Expression Segmentation}. In \bibinfo{booktitle}{\emph{European Conference on Computer Vision}}. Springer, \bibinfo{pages}{323--340}.
\newblock


\bibitem[Chen et~al\mbox{.}(2024b)]%
        {chen2024internvl}
\bibfield{author}{\bibinfo{person}{Zhe Chen}, \bibinfo{person}{Jiannan Wu}, \bibinfo{person}{Wenhai Wang}, \bibinfo{person}{Weijie Su}, \bibinfo{person}{Guo Chen}, \bibinfo{person}{Sen Xing}, \bibinfo{person}{Muyan Zhong}, \bibinfo{person}{Qinglong Zhang}, \bibinfo{person}{Xizhou Zhu}, \bibinfo{person}{Lewei Lu}, {et~al\mbox{.}}} \bibinfo{year}{2024}\natexlab{b}.
\newblock \showarticletitle{Internvl: Scaling up vision foundation models and aligning for generic visual-linguistic tasks}. In \bibinfo{booktitle}{\emph{Proceedings of the IEEE/CVF conference on computer vision and pattern recognition}}. \bibinfo{pages}{24185--24198}.
\newblock


\bibitem[Chiang et~al\mbox{.}(2023)]%
        {chiang2023vicuna}
\bibfield{author}{\bibinfo{person}{Wei-Lin Chiang}, \bibinfo{person}{Zhuohan Li}, \bibinfo{person}{Zi Lin}, \bibinfo{person}{Ying Sheng}, \bibinfo{person}{Zhanghao Wu}, \bibinfo{person}{Hao Zhang}, \bibinfo{person}{Lianmin Zheng}, \bibinfo{person}{Siyuan Zhuang}, \bibinfo{person}{Yonghao Zhuang}, \bibinfo{person}{Joseph~E Gonzalez}, {et~al\mbox{.}}} \bibinfo{year}{2023}\natexlab{}.
\newblock \showarticletitle{Vicuna: An open-source chatbot impressing gpt-4 with 90\%* chatgpt quality}.
\newblock \bibinfo{journal}{\emph{See https://vicuna. lmsys. org (accessed 14 April 2023)}} \bibinfo{volume}{2}, \bibinfo{number}{3} (\bibinfo{year}{2023}), \bibinfo{pages}{6}.
\newblock


\bibitem[Ding et~al\mbox{.}(2021)]%
        {ding2021vision}
\bibfield{author}{\bibinfo{person}{Henghui Ding}, \bibinfo{person}{Chang Liu}, \bibinfo{person}{Suchen Wang}, {and} \bibinfo{person}{Xudong Jiang}.} \bibinfo{year}{2021}\natexlab{}.
\newblock \showarticletitle{Vision-language transformer and query generation for referring segmentation}. In \bibinfo{booktitle}{\emph{Proceedings of the IEEE/CVF International Conference on Computer Vision}}. \bibinfo{pages}{16321--16330}.
\newblock


\bibitem[Dong et~al\mbox{.}(2024)]%
        {dong2024internlm}
\bibfield{author}{\bibinfo{person}{Xiaoyi Dong}, \bibinfo{person}{Pan Zhang}, \bibinfo{person}{Yuhang Zang}, \bibinfo{person}{Yuhang Cao}, \bibinfo{person}{Bin Wang}, \bibinfo{person}{Linke Ouyang}, \bibinfo{person}{Songyang Zhang}, \bibinfo{person}{Haodong Duan}, \bibinfo{person}{Wenwei Zhang}, \bibinfo{person}{Yining Li}, {et~al\mbox{.}}} \bibinfo{year}{2024}\natexlab{}.
\newblock \showarticletitle{Internlm-xcomposer2-4khd: A pioneering large vision-language model handling resolutions from 336 pixels to 4k hd}.
\newblock \bibinfo{journal}{\emph{arXiv preprint arXiv:2404.06512}} (\bibinfo{year}{2024}).
\newblock


\bibitem[Dosovitskiy(2020)]%
        {dosovitskiy2020image}
\bibfield{author}{\bibinfo{person}{Alexey Dosovitskiy}.} \bibinfo{year}{2020}\natexlab{}.
\newblock \showarticletitle{An image is worth 16x16 words: Transformers for image recognition at scale}.
\newblock \bibinfo{journal}{\emph{arXiv preprint arXiv:2010.11929}} (\bibinfo{year}{2020}).
\newblock


\bibitem[Guo et~al\mbox{.}(2025b)]%
        {guo2025deepseek}
\bibfield{author}{\bibinfo{person}{Daya Guo}, \bibinfo{person}{Dejian Yang}, \bibinfo{person}{Haowei Zhang}, \bibinfo{person}{Junxiao Song}, \bibinfo{person}{Ruoyu Zhang}, \bibinfo{person}{Runxin Xu}, \bibinfo{person}{Qihao Zhu}, \bibinfo{person}{Shirong Ma}, \bibinfo{person}{Peiyi Wang}, \bibinfo{person}{Xiao Bi}, {et~al\mbox{.}}} \bibinfo{year}{2025}\natexlab{b}.
\newblock \showarticletitle{Deepseek-r1: Incentivizing reasoning capability in llms via reinforcement learning}.
\newblock \bibinfo{journal}{\emph{arXiv preprint arXiv:2501.12948}} (\bibinfo{year}{2025}).
\newblock


\bibitem[Guo et~al\mbox{.}(2025a)]%
        {guo2025llava}
\bibfield{author}{\bibinfo{person}{Zonghao Guo}, \bibinfo{person}{Ruyi Xu}, \bibinfo{person}{Yuan Yao}, \bibinfo{person}{Junbo Cui}, \bibinfo{person}{Zanlin Ni}, \bibinfo{person}{Chunjiang Ge}, \bibinfo{person}{Tat-Seng Chua}, \bibinfo{person}{Zhiyuan Liu}, {and} \bibinfo{person}{Gao Huang}.} \bibinfo{year}{2025}\natexlab{a}.
\newblock \showarticletitle{Llava-uhd: an lmm perceiving any aspect ratio and high-resolution images}. In \bibinfo{booktitle}{\emph{European Conference on Computer Vision}}. Springer, \bibinfo{pages}{390--406}.
\newblock


\bibitem[Gupta et~al\mbox{.}(2019)]%
        {gupta2019lvis}
\bibfield{author}{\bibinfo{person}{Agrim Gupta}, \bibinfo{person}{Piotr Dollar}, {and} \bibinfo{person}{Ross Girshick}.} \bibinfo{year}{2019}\natexlab{}.
\newblock \showarticletitle{Lvis: A dataset for large vocabulary instance segmentation}. In \bibinfo{booktitle}{\emph{Proceedings of the IEEE/CVF conference on computer vision and pattern recognition}}. \bibinfo{pages}{5356--5364}.
\newblock


\bibitem[He et~al\mbox{.}(2022)]%
        {he2022partimagenet}
\bibfield{author}{\bibinfo{person}{Ju He}, \bibinfo{person}{Shuo Yang}, \bibinfo{person}{Shaokang Yang}, \bibinfo{person}{Adam Kortylewski}, \bibinfo{person}{Xiaoding Yuan}, \bibinfo{person}{Jie-Neng Chen}, \bibinfo{person}{Shuai Liu}, \bibinfo{person}{Cheng Yang}, \bibinfo{person}{Qihang Yu}, {and} \bibinfo{person}{Alan Yuille}.} \bibinfo{year}{2022}\natexlab{}.
\newblock \showarticletitle{Partimagenet: A large, high-quality dataset of parts}. In \bibinfo{booktitle}{\emph{European Conference on Computer Vision}}. Springer, \bibinfo{pages}{128--145}.
\newblock


\bibitem[Hu et~al\mbox{.}(2024)]%
        {hu2024mplug}
\bibfield{author}{\bibinfo{person}{Anwen Hu}, \bibinfo{person}{Haiyang Xu}, \bibinfo{person}{Jiabo Ye}, \bibinfo{person}{Ming Yan}, \bibinfo{person}{Liang Zhang}, \bibinfo{person}{Bo Zhang}, \bibinfo{person}{Chen Li}, \bibinfo{person}{Ji Zhang}, \bibinfo{person}{Qin Jin}, \bibinfo{person}{Fei Huang}, {et~al\mbox{.}}} \bibinfo{year}{2024}\natexlab{}.
\newblock \showarticletitle{mplug-docowl 1.5: Unified structure learning for ocr-free document understanding}.
\newblock \bibinfo{journal}{\emph{arXiv preprint arXiv:2403.12895}} (\bibinfo{year}{2024}).
\newblock


\bibitem[Hu et~al\mbox{.}(2016)]%
        {hu2016segmentation}
\bibfield{author}{\bibinfo{person}{Ronghang Hu}, \bibinfo{person}{Marcus Rohrbach}, {and} \bibinfo{person}{Trevor Darrell}.} \bibinfo{year}{2016}\natexlab{}.
\newblock \showarticletitle{Segmentation from natural language expressions}. In \bibinfo{booktitle}{\emph{Computer Vision--ECCV 2016: 14th European Conference, Amsterdam, The Netherlands, October 11--14, 2016, Proceedings, Part I 14}}. Springer, \bibinfo{pages}{108--124}.
\newblock


\bibitem[Huang et~al\mbox{.}(2024)]%
        {huang2024mini}
\bibfield{author}{\bibinfo{person}{Mingxin Huang}, \bibinfo{person}{Yuliang Liu}, \bibinfo{person}{Dingkang Liang}, \bibinfo{person}{Lianwen Jin}, {and} \bibinfo{person}{Xiang Bai}.} \bibinfo{year}{2024}\natexlab{}.
\newblock \showarticletitle{Mini-monkey: Alleviating the semantic sawtooth effect for lightweight mllms via complementary image pyramid}.
\newblock \bibinfo{journal}{\emph{arXiv preprint arXiv:2408.02034}} (\bibinfo{year}{2024}).
\newblock


\bibitem[Jang et~al\mbox{.}(2025)]%
        {jang2025mmr}
\bibfield{author}{\bibinfo{person}{Donggon Jang}, \bibinfo{person}{Yucheol Cho}, \bibinfo{person}{Suin Lee}, \bibinfo{person}{Taehyeon Kim}, {and} \bibinfo{person}{Dae-Shik Kim}.} \bibinfo{year}{2025}\natexlab{}.
\newblock \showarticletitle{MMR: A Large-scale Benchmark Dataset for Multi-target and Multi-granularity Reasoning Segmentation}.
\newblock \bibinfo{journal}{\emph{arXiv preprint arXiv:2503.13881}} (\bibinfo{year}{2025}).
\newblock


\bibitem[Jiang et~al\mbox{.}(2024)]%
        {jiang2024mixtral}
\bibfield{author}{\bibinfo{person}{Albert~Q Jiang}, \bibinfo{person}{Alexandre Sablayrolles}, \bibinfo{person}{Antoine Roux}, \bibinfo{person}{Arthur Mensch}, \bibinfo{person}{Blanche Savary}, \bibinfo{person}{Chris Bamford}, \bibinfo{person}{Devendra~Singh Chaplot}, \bibinfo{person}{Diego de~las Casas}, \bibinfo{person}{Emma~Bou Hanna}, \bibinfo{person}{Florian Bressand}, {et~al\mbox{.}}} \bibinfo{year}{2024}\natexlab{}.
\newblock \showarticletitle{Mixtral of experts}.
\newblock \bibinfo{journal}{\emph{arXiv preprint arXiv:2401.04088}} (\bibinfo{year}{2024}).
\newblock


\bibitem[Kazemzadeh et~al\mbox{.}(2014)]%
        {kazemzadeh2014referitgame}
\bibfield{author}{\bibinfo{person}{Sahar Kazemzadeh}, \bibinfo{person}{Vicente Ordonez}, \bibinfo{person}{Mark Matten}, {and} \bibinfo{person}{Tamara Berg}.} \bibinfo{year}{2014}\natexlab{}.
\newblock \showarticletitle{Referitgame: Referring to objects in photographs of natural scenes}. In \bibinfo{booktitle}{\emph{Proceedings of the 2014 conference on empirical methods in natural language processing (EMNLP)}}. \bibinfo{pages}{787--798}.
\newblock


\bibitem[Kirillov et~al\mbox{.}(2023)]%
        {kirillov2023segment}
\bibfield{author}{\bibinfo{person}{Alexander Kirillov}, \bibinfo{person}{Eric Mintun}, \bibinfo{person}{Nikhila Ravi}, \bibinfo{person}{Hanzi Mao}, \bibinfo{person}{Chloe Rolland}, \bibinfo{person}{Laura Gustafson}, \bibinfo{person}{Tete Xiao}, \bibinfo{person}{Spencer Whitehead}, \bibinfo{person}{Alexander~C Berg}, \bibinfo{person}{Wan-Yen Lo}, {et~al\mbox{.}}} \bibinfo{year}{2023}\natexlab{}.
\newblock \showarticletitle{Segment anything}. In \bibinfo{booktitle}{\emph{Proceedings of the IEEE/CVF International Conference on Computer Vision}}. \bibinfo{pages}{4015--4026}.
\newblock


\bibitem[Lai et~al\mbox{.}(2024)]%
        {lai2024lisa}
\bibfield{author}{\bibinfo{person}{Xin Lai}, \bibinfo{person}{Zhuotao Tian}, \bibinfo{person}{Yukang Chen}, \bibinfo{person}{Yanwei Li}, \bibinfo{person}{Yuhui Yuan}, \bibinfo{person}{Shu Liu}, {and} \bibinfo{person}{Jiaya Jia}.} \bibinfo{year}{2024}\natexlab{}.
\newblock \showarticletitle{Lisa: Reasoning segmentation via large language model}. In \bibinfo{booktitle}{\emph{Proceedings of the IEEE/CVF Conference on Computer Vision and Pattern Recognition}}. \bibinfo{pages}{9579--9589}.
\newblock


\bibitem[Li et~al\mbox{.}(2023)]%
        {li2023blip}
\bibfield{author}{\bibinfo{person}{Junnan Li}, \bibinfo{person}{Dongxu Li}, \bibinfo{person}{Silvio Savarese}, {and} \bibinfo{person}{Steven Hoi}.} \bibinfo{year}{2023}\natexlab{}.
\newblock \showarticletitle{Blip-2: Bootstrapping language-image pre-training with frozen image encoders and large language models}. In \bibinfo{booktitle}{\emph{International conference on machine learning}}. PMLR, \bibinfo{pages}{19730--19742}.
\newblock


\bibitem[Li et~al\mbox{.}(2024c)]%
        {li2024tokenpacker}
\bibfield{author}{\bibinfo{person}{Wentong Li}, \bibinfo{person}{Yuqian Yuan}, \bibinfo{person}{Jian Liu}, \bibinfo{person}{Dongqi Tang}, \bibinfo{person}{Song Wang}, \bibinfo{person}{Jie Qin}, \bibinfo{person}{Jianke Zhu}, {and} \bibinfo{person}{Lei Zhang}.} \bibinfo{year}{2024}\natexlab{c}.
\newblock \showarticletitle{Tokenpacker: Efficient visual projector for multimodal llm}.
\newblock \bibinfo{journal}{\emph{arXiv preprint arXiv:2407.02392}} (\bibinfo{year}{2024}).
\newblock


\bibitem[Li et~al\mbox{.}(2024b)]%
        {li2024omg}
\bibfield{author}{\bibinfo{person}{Xiangtai Li}, \bibinfo{person}{Haobo Yuan}, \bibinfo{person}{Wei Li}, \bibinfo{person}{Henghui Ding}, \bibinfo{person}{Size Wu}, \bibinfo{person}{Wenwei Zhang}, \bibinfo{person}{Yining Li}, \bibinfo{person}{Kai Chen}, {and} \bibinfo{person}{Chen~Change Loy}.} \bibinfo{year}{2024}\natexlab{b}.
\newblock \showarticletitle{Omg-seg: Is one model good enough for all segmentation?}. In \bibinfo{booktitle}{\emph{Proceedings of the IEEE/CVF conference on computer vision and pattern recognition}}. \bibinfo{pages}{27948--27959}.
\newblock


\bibitem[Li et~al\mbox{.}(2024d)]%
        {li2024mini}
\bibfield{author}{\bibinfo{person}{Yanwei Li}, \bibinfo{person}{Yuechen Zhang}, \bibinfo{person}{Chengyao Wang}, \bibinfo{person}{Zhisheng Zhong}, \bibinfo{person}{Yixin Chen}, \bibinfo{person}{Ruihang Chu}, \bibinfo{person}{Shaoteng Liu}, {and} \bibinfo{person}{Jiaya Jia}.} \bibinfo{year}{2024}\natexlab{d}.
\newblock \showarticletitle{Mini-gemini: Mining the potential of multi-modality vision language models}.
\newblock \bibinfo{journal}{\emph{arXiv preprint arXiv:2403.18814}} (\bibinfo{year}{2024}).
\newblock


\bibitem[Li et~al\mbox{.}(2024a)]%
        {li2024monkey}
\bibfield{author}{\bibinfo{person}{Zhang Li}, \bibinfo{person}{Biao Yang}, \bibinfo{person}{Qiang Liu}, \bibinfo{person}{Zhiyin Ma}, \bibinfo{person}{Shuo Zhang}, \bibinfo{person}{Jingxu Yang}, \bibinfo{person}{Yabo Sun}, \bibinfo{person}{Yuliang Liu}, {and} \bibinfo{person}{Xiang Bai}.} \bibinfo{year}{2024}\natexlab{a}.
\newblock \showarticletitle{Monkey: Image resolution and text label are important things for large multi-modal models}. In \bibinfo{booktitle}{\emph{Proceedings of the IEEE/CVF Conference on Computer Vision and Pattern Recognition}}. \bibinfo{pages}{26763--26773}.
\newblock


\bibitem[Liang et~al\mbox{.}(2023)]%
        {liang2023open}
\bibfield{author}{\bibinfo{person}{Feng Liang}, \bibinfo{person}{Bichen Wu}, \bibinfo{person}{Xiaoliang Dai}, \bibinfo{person}{Kunpeng Li}, \bibinfo{person}{Yinan Zhao}, \bibinfo{person}{Hang Zhang}, \bibinfo{person}{Peizhao Zhang}, \bibinfo{person}{Peter Vajda}, {and} \bibinfo{person}{Diana Marculescu}.} \bibinfo{year}{2023}\natexlab{}.
\newblock \showarticletitle{Open-vocabulary semantic segmentation with mask-adapted clip}. In \bibinfo{booktitle}{\emph{Proceedings of the IEEE/CVF Conference on Computer Vision and Pattern Recognition}}. \bibinfo{pages}{7061--7070}.
\newblock


\bibitem[Lin et~al\mbox{.}(2023)]%
        {lin2023sphinx}
\bibfield{author}{\bibinfo{person}{Ziyi Lin}, \bibinfo{person}{Chris Liu}, \bibinfo{person}{Renrui Zhang}, \bibinfo{person}{Peng Gao}, \bibinfo{person}{Longtian Qiu}, \bibinfo{person}{Han Xiao}, \bibinfo{person}{Han Qiu}, \bibinfo{person}{Chen Lin}, \bibinfo{person}{Wenqi Shao}, \bibinfo{person}{Keqin Chen}, {et~al\mbox{.}}} \bibinfo{year}{2023}\natexlab{}.
\newblock \showarticletitle{Sphinx: The joint mixing of weights, tasks, and visual embeddings for multi-modal large language models}.
\newblock \bibinfo{journal}{\emph{arXiv preprint arXiv:2311.07575}} (\bibinfo{year}{2023}).
\newblock


\bibitem[Liu et~al\mbox{.}(2024b)]%
        {liu2024deepseek}
\bibfield{author}{\bibinfo{person}{Aixin Liu}, \bibinfo{person}{Bei Feng}, \bibinfo{person}{Bing Xue}, \bibinfo{person}{Bingxuan Wang}, \bibinfo{person}{Bochao Wu}, \bibinfo{person}{Chengda Lu}, \bibinfo{person}{Chenggang Zhao}, \bibinfo{person}{Chengqi Deng}, \bibinfo{person}{Chenyu Zhang}, \bibinfo{person}{Chong Ruan}, {et~al\mbox{.}}} \bibinfo{year}{2024}\natexlab{b}.
\newblock \showarticletitle{Deepseek-v3 technical report}.
\newblock \bibinfo{journal}{\emph{arXiv preprint arXiv:2412.19437}} (\bibinfo{year}{2024}).
\newblock


\bibitem[Liu et~al\mbox{.}(2023a)]%
        {liu2023gres}
\bibfield{author}{\bibinfo{person}{Chang Liu}, \bibinfo{person}{Henghui Ding}, {and} \bibinfo{person}{Xudong Jiang}.} \bibinfo{year}{2023}\natexlab{a}.
\newblock \showarticletitle{Gres: Generalized referring expression segmentation}. In \bibinfo{booktitle}{\emph{Proceedings of the IEEE/CVF conference on computer vision and pattern recognition}}. \bibinfo{pages}{23592--23601}.
\newblock


\bibitem[Liu et~al\mbox{.}(2024c)]%
        {liu2024llava}
\bibfield{author}{\bibinfo{person}{Haotian Liu}, \bibinfo{person}{Chunyuan Li}, \bibinfo{person}{Yuheng Li}, \bibinfo{person}{Bo Li}, \bibinfo{person}{Yuanhan Zhang}, \bibinfo{person}{Sheng Shen}, {and} \bibinfo{person}{Yong~Jae Lee}.} \bibinfo{year}{2024}\natexlab{c}.
\newblock \bibinfo{title}{Llava-next: Improved reasoning, ocr, and world knowledge}.
\newblock


\bibitem[Liu et~al\mbox{.}(2023b)]%
        {liu2023visual}
\bibfield{author}{\bibinfo{person}{Haotian Liu}, \bibinfo{person}{Chunyuan Li}, \bibinfo{person}{Qingyang Wu}, {and} \bibinfo{person}{Yong~Jae Lee}.} \bibinfo{year}{2023}\natexlab{b}.
\newblock \showarticletitle{Visual Instruction Tuning}.
\newblock \bibinfo{journal}{\emph{arXiv preprint arXiv:2304.08485}} (\bibinfo{year}{2023}).
\newblock


\bibitem[Liu et~al\mbox{.}(2025b)]%
        {liu2025grounding}
\bibfield{author}{\bibinfo{person}{Shilong Liu}, \bibinfo{person}{Zhaoyang Zeng}, \bibinfo{person}{Tianhe Ren}, \bibinfo{person}{Feng Li}, \bibinfo{person}{Hao Zhang}, \bibinfo{person}{Jie Yang}, \bibinfo{person}{Qing Jiang}, \bibinfo{person}{Chunyuan Li}, \bibinfo{person}{Jianwei Yang}, \bibinfo{person}{Hang Su}, {et~al\mbox{.}}} \bibinfo{year}{2025}\natexlab{b}.
\newblock \showarticletitle{Grounding dino: Marrying dino with grounded pre-training for open-set object detection}. In \bibinfo{booktitle}{\emph{European Conference on Computer Vision}}. Springer, \bibinfo{pages}{38--55}.
\newblock


\bibitem[Liu et~al\mbox{.}(2025a)]%
        {liu2025seg}
\bibfield{author}{\bibinfo{person}{Yuqi Liu}, \bibinfo{person}{Bohao Peng}, \bibinfo{person}{Zhisheng Zhong}, \bibinfo{person}{Zihao Yue}, \bibinfo{person}{Fanbin Lu}, \bibinfo{person}{Bei Yu}, {and} \bibinfo{person}{Jiaya Jia}.} \bibinfo{year}{2025}\natexlab{a}.
\newblock \showarticletitle{Seg-zero: Reasoning-chain guided segmentation via cognitive reinforcement}.
\newblock \bibinfo{journal}{\emph{arXiv preprint arXiv:2503.06520}} (\bibinfo{year}{2025}).
\newblock


\bibitem[Liu et~al\mbox{.}(2024a)]%
        {liu2024oryx}
\bibfield{author}{\bibinfo{person}{Zuyan Liu}, \bibinfo{person}{Yuhao Dong}, \bibinfo{person}{Ziwei Liu}, \bibinfo{person}{Winston Hu}, \bibinfo{person}{Jiwen Lu}, {and} \bibinfo{person}{Yongming Rao}.} \bibinfo{year}{2024}\natexlab{a}.
\newblock \showarticletitle{Oryx mllm: On-demand spatial-temporal understanding at arbitrary resolution}.
\newblock \bibinfo{journal}{\emph{arXiv preprint arXiv:2409.12961}} (\bibinfo{year}{2024}).
\newblock


\bibitem[Liu et~al\mbox{.}(2021a)]%
        {liu2021swin}
\bibfield{author}{\bibinfo{person}{Ze Liu}, \bibinfo{person}{Yutong Lin}, \bibinfo{person}{Yue Cao}, \bibinfo{person}{Han Hu}, \bibinfo{person}{Yixuan Wei}, \bibinfo{person}{Zheng Zhang}, \bibinfo{person}{Stephen Lin}, {and} \bibinfo{person}{Baining Guo}.} \bibinfo{year}{2021}\natexlab{a}.
\newblock \showarticletitle{Swin transformer: Hierarchical vision transformer using shifted windows}. In \bibinfo{booktitle}{\emph{Proceedings of the IEEE/CVF international conference on computer vision}}. \bibinfo{pages}{10012--10022}.
\newblock


\bibitem[Liu et~al\mbox{.}(2021b)]%
        {liu2021post}
\bibfield{author}{\bibinfo{person}{Zhenhua Liu}, \bibinfo{person}{Yunhe Wang}, \bibinfo{person}{Kai Han}, \bibinfo{person}{Wei Zhang}, \bibinfo{person}{Siwei Ma}, {and} \bibinfo{person}{Wen Gao}.} \bibinfo{year}{2021}\natexlab{b}.
\newblock \showarticletitle{Post-training quantization for vision transformer}.
\newblock \bibinfo{journal}{\emph{Advances in Neural Information Processing Systems}}  \bibinfo{volume}{34} (\bibinfo{year}{2021}), \bibinfo{pages}{28092--28103}.
\newblock


\bibitem[Loshchilov(2017)]%
        {loshchilov2017decoupled}
\bibfield{author}{\bibinfo{person}{I Loshchilov}.} \bibinfo{year}{2017}\natexlab{}.
\newblock \showarticletitle{Decoupled weight decay regularization}.
\newblock \bibinfo{journal}{\emph{arXiv preprint arXiv:1711.05101}} (\bibinfo{year}{2017}).
\newblock


\bibitem[Lu et~al\mbox{.}(2024)]%
        {lu2024deepseek}
\bibfield{author}{\bibinfo{person}{Haoyu Lu}, \bibinfo{person}{Wen Liu}, \bibinfo{person}{Bo Zhang}, \bibinfo{person}{Bingxuan Wang}, \bibinfo{person}{Kai Dong}, \bibinfo{person}{Bo Liu}, \bibinfo{person}{Jingxiang Sun}, \bibinfo{person}{Tongzheng Ren}, \bibinfo{person}{Zhuoshu Li}, \bibinfo{person}{Hao Yang}, {et~al\mbox{.}}} \bibinfo{year}{2024}\natexlab{}.
\newblock \showarticletitle{Deepseek-vl: towards real-world vision-language understanding}.
\newblock \bibinfo{journal}{\emph{arXiv preprint arXiv:2403.05525}} (\bibinfo{year}{2024}).
\newblock


\bibitem[Luo et~al\mbox{.}(2020)]%
        {luo2020multi}
\bibfield{author}{\bibinfo{person}{Gen Luo}, \bibinfo{person}{Yiyi Zhou}, \bibinfo{person}{Xiaoshuai Sun}, \bibinfo{person}{Liujuan Cao}, \bibinfo{person}{Chenglin Wu}, \bibinfo{person}{Cheng Deng}, {and} \bibinfo{person}{Rongrong Ji}.} \bibinfo{year}{2020}\natexlab{}.
\newblock \showarticletitle{Multi-task collaborative network for joint referring expression comprehension and segmentation}. In \bibinfo{booktitle}{\emph{Proceedings of the IEEE/CVF Conference on computer vision and pattern recognition}}. \bibinfo{pages}{10034--10043}.
\newblock


\bibitem[Ma et~al\mbox{.}(2024)]%
        {ma2024inf}
\bibfield{author}{\bibinfo{person}{Yiwei Ma}, \bibinfo{person}{Zhibin Wang}, \bibinfo{person}{Xiaoshuai Sun}, \bibinfo{person}{Weihuang Lin}, \bibinfo{person}{Qiang Zhou}, \bibinfo{person}{Jiayi Ji}, {and} \bibinfo{person}{Rongrong Ji}.} \bibinfo{year}{2024}\natexlab{}.
\newblock \showarticletitle{INF-LLaVA: Dual-perspective Perception for High-Resolution Multimodal Large Language Model}.
\newblock \bibinfo{journal}{\emph{arXiv preprint arXiv:2407.16198}} (\bibinfo{year}{2024}).
\newblock


\bibitem[Mao et~al\mbox{.}(2016)]%
        {mao2016generation}
\bibfield{author}{\bibinfo{person}{Junhua Mao}, \bibinfo{person}{Jonathan Huang}, \bibinfo{person}{Alexander Toshev}, \bibinfo{person}{Oana Camburu}, \bibinfo{person}{Alan~L Yuille}, {and} \bibinfo{person}{Kevin Murphy}.} \bibinfo{year}{2016}\natexlab{}.
\newblock \showarticletitle{Generation and comprehension of unambiguous object descriptions}. In \bibinfo{booktitle}{\emph{Proceedings of the IEEE conference on computer vision and pattern recognition}}. \bibinfo{pages}{11--20}.
\newblock


\bibitem[Meta(2024)]%
        {meta2024introducing}
\bibfield{author}{\bibinfo{person}{AI Meta}.} \bibinfo{year}{2024}\natexlab{}.
\newblock \showarticletitle{Introducing meta llama 3: The most capable openly available llm to date}.
\newblock \bibinfo{journal}{\emph{Meta AI}} (\bibinfo{year}{2024}).
\newblock


\bibitem[OpenAI(2023)]%
        {openai2023gpt}
\bibfield{author}{\bibinfo{person}{R OpenAI}.} \bibinfo{year}{2023}\natexlab{}.
\newblock \showarticletitle{Gpt-4 technical report. arxiv 2303.08774}.
\newblock \bibinfo{journal}{\emph{View in Article}} \bibinfo{volume}{2}, \bibinfo{number}{5} (\bibinfo{year}{2023}).
\newblock


\bibitem[Oquab et~al\mbox{.}(2023)]%
        {oquab2023dinov2}
\bibfield{author}{\bibinfo{person}{Maxime Oquab}, \bibinfo{person}{Timoth{\'e}e Darcet}, \bibinfo{person}{Th{\'e}o Moutakanni}, \bibinfo{person}{Huy Vo}, \bibinfo{person}{Marc Szafraniec}, \bibinfo{person}{Vasil Khalidov}, \bibinfo{person}{Pierre Fernandez}, \bibinfo{person}{Daniel Haziza}, \bibinfo{person}{Francisco Massa}, \bibinfo{person}{Alaaeldin El-Nouby}, {et~al\mbox{.}}} \bibinfo{year}{2023}\natexlab{}.
\newblock \showarticletitle{Dinov2: Learning robust visual features without supervision}.
\newblock \bibinfo{journal}{\emph{arXiv preprint arXiv:2304.07193}} (\bibinfo{year}{2023}).
\newblock


\bibitem[Ramanathan et~al\mbox{.}(2023)]%
        {ramanathan2023paco}
\bibfield{author}{\bibinfo{person}{Vignesh Ramanathan}, \bibinfo{person}{Anmol Kalia}, \bibinfo{person}{Vladan Petrovic}, \bibinfo{person}{Yi Wen}, \bibinfo{person}{Baixue Zheng}, \bibinfo{person}{Baishan Guo}, \bibinfo{person}{Rui Wang}, \bibinfo{person}{Aaron Marquez}, \bibinfo{person}{Rama Kovvuri}, \bibinfo{person}{Abhishek Kadian}, {et~al\mbox{.}}} \bibinfo{year}{2023}\natexlab{}.
\newblock \showarticletitle{Paco: Parts and attributes of common objects}. In \bibinfo{booktitle}{\emph{Proceedings of the IEEE/CVF Conference on Computer Vision and Pattern Recognition}}. \bibinfo{pages}{7141--7151}.
\newblock


\bibitem[Rasheed et~al\mbox{.}(2024)]%
        {rasheed2024glamm}
\bibfield{author}{\bibinfo{person}{Hanoona Rasheed}, \bibinfo{person}{Muhammad Maaz}, \bibinfo{person}{Sahal Shaji}, \bibinfo{person}{Abdelrahman Shaker}, \bibinfo{person}{Salman Khan}, \bibinfo{person}{Hisham Cholakkal}, \bibinfo{person}{Rao~M Anwer}, \bibinfo{person}{Eric Xing}, \bibinfo{person}{Ming-Hsuan Yang}, {and} \bibinfo{person}{Fahad~S Khan}.} \bibinfo{year}{2024}\natexlab{}.
\newblock \showarticletitle{Glamm: Pixel grounding large multimodal model}. In \bibinfo{booktitle}{\emph{Proceedings of the IEEE/CVF Conference on Computer Vision and Pattern Recognition}}. \bibinfo{pages}{13009--13018}.
\newblock


\bibitem[Ren et~al\mbox{.}(2024)]%
        {ren2024pixellm}
\bibfield{author}{\bibinfo{person}{Zhongwei Ren}, \bibinfo{person}{Zhicheng Huang}, \bibinfo{person}{Yunchao Wei}, \bibinfo{person}{Yao Zhao}, \bibinfo{person}{Dongmei Fu}, \bibinfo{person}{Jiashi Feng}, {and} \bibinfo{person}{Xiaojie Jin}.} \bibinfo{year}{2024}\natexlab{}.
\newblock \showarticletitle{Pixellm: Pixel reasoning with large multimodal model}. In \bibinfo{booktitle}{\emph{Proceedings of the IEEE/CVF Conference on Computer Vision and Pattern Recognition}}. \bibinfo{pages}{26374--26383}.
\newblock


\bibitem[Team(2023)]%
        {team2023internlm}
\bibfield{author}{\bibinfo{person}{InternLM Team}.} \bibinfo{year}{2023}\natexlab{}.
\newblock \bibinfo{title}{Internlm: A multilingual language model with progressively enhanced capabilities}.
\newblock


\bibitem[Touvron et~al\mbox{.}(2023)]%
        {touvron2023llama}
\bibfield{author}{\bibinfo{person}{Hugo Touvron}, \bibinfo{person}{Thibaut Lavril}, \bibinfo{person}{Gautier Izacard}, \bibinfo{person}{Xavier Martinet}, \bibinfo{person}{Marie-Anne Lachaux}, \bibinfo{person}{Timoth{\'e}e Lacroix}, \bibinfo{person}{Baptiste Rozi{\`e}re}, \bibinfo{person}{Naman Goyal}, \bibinfo{person}{Eric Hambro}, \bibinfo{person}{Faisal Azhar}, {et~al\mbox{.}}} \bibinfo{year}{2023}\natexlab{}.
\newblock \showarticletitle{LLaMA: open and efficient foundation language models. arXiv}.
\newblock \bibinfo{journal}{\emph{arXiv preprint arXiv:2302.13971}} (\bibinfo{year}{2023}).
\newblock


\bibitem[Tschannen et~al\mbox{.}(2025)]%
        {tschannen2025siglip}
\bibfield{author}{\bibinfo{person}{Michael Tschannen}, \bibinfo{person}{Alexey Gritsenko}, \bibinfo{person}{Xiao Wang}, \bibinfo{person}{Muhammad~Ferjad Naeem}, \bibinfo{person}{Ibrahim Alabdulmohsin}, \bibinfo{person}{Nikhil Parthasarathy}, \bibinfo{person}{Talfan Evans}, \bibinfo{person}{Lucas Beyer}, \bibinfo{person}{Ye Xia}, \bibinfo{person}{Basil Mustafa}, {et~al\mbox{.}}} \bibinfo{year}{2025}\natexlab{}.
\newblock \showarticletitle{Siglip 2: Multilingual vision-language encoders with improved semantic understanding, localization, and dense features}.
\newblock \bibinfo{journal}{\emph{arXiv preprint arXiv:2502.14786}} (\bibinfo{year}{2025}).
\newblock


\bibitem[Wang and Ke(2024)]%
        {wang2024llm}
\bibfield{author}{\bibinfo{person}{Junchi Wang} {and} \bibinfo{person}{Lei Ke}.} \bibinfo{year}{2024}\natexlab{}.
\newblock \showarticletitle{LLM-Seg: Bridging Image Segmentation and Large Language Model Reasoning}. In \bibinfo{booktitle}{\emph{Proceedings of the IEEE/CVF Conference on Computer Vision and Pattern Recognition}}. \bibinfo{pages}{1765--1774}.
\newblock


\bibitem[Wang et~al\mbox{.}(2024a)]%
        {wang2024qwen2}
\bibfield{author}{\bibinfo{person}{Peng Wang}, \bibinfo{person}{Shuai Bai}, \bibinfo{person}{Sinan Tan}, \bibinfo{person}{Shijie Wang}, \bibinfo{person}{Zhihao Fan}, \bibinfo{person}{Jinze Bai}, \bibinfo{person}{Keqin Chen}, \bibinfo{person}{Xuejing Liu}, \bibinfo{person}{Jialin Wang}, \bibinfo{person}{Wenbin Ge}, {et~al\mbox{.}}} \bibinfo{year}{2024}\natexlab{a}.
\newblock \showarticletitle{Qwen2-vl: Enhancing vision-language model's perception of the world at any resolution}.
\newblock \bibinfo{journal}{\emph{arXiv preprint arXiv:2409.12191}} (\bibinfo{year}{2024}).
\newblock


\bibitem[Wang et~al\mbox{.}(2023)]%
        {wang2023visionllm}
\bibfield{author}{\bibinfo{person}{Wenhai Wang}, \bibinfo{person}{Zhe Chen}, \bibinfo{person}{Xiaokang Chen}, \bibinfo{person}{Jiannan Wu}, \bibinfo{person}{Xizhou Zhu}, \bibinfo{person}{Gang Zeng}, \bibinfo{person}{Ping Luo}, \bibinfo{person}{Tong Lu}, \bibinfo{person}{Jie Zhou}, \bibinfo{person}{Yu Qiao}, {et~al\mbox{.}}} \bibinfo{year}{2023}\natexlab{}.
\newblock \showarticletitle{Visionllm: Large language model is also an open-ended decoder for vision-centric tasks}.
\newblock \bibinfo{journal}{\emph{Advances in Neural Information Processing Systems}}  \bibinfo{volume}{36} (\bibinfo{year}{2023}), \bibinfo{pages}{61501--61513}.
\newblock


\bibitem[Wang et~al\mbox{.}(2024b)]%
        {wang2024segllm}
\bibfield{author}{\bibinfo{person}{XuDong Wang}, \bibinfo{person}{Shaolun Zhang}, \bibinfo{person}{Shufan Li}, \bibinfo{person}{Konstantinos Kallidromitis}, \bibinfo{person}{Kehan Li}, \bibinfo{person}{Yusuke Kato}, \bibinfo{person}{Kazuki Kozuka}, {and} \bibinfo{person}{Trevor Darrell}.} \bibinfo{year}{2024}\natexlab{b}.
\newblock \showarticletitle{SegLLM: Multi-round Reasoning Segmentation}.
\newblock \bibinfo{journal}{\emph{arXiv preprint arXiv:2410.18923}} (\bibinfo{year}{2024}).
\newblock


\bibitem[Wang et~al\mbox{.}(2022)]%
        {wang2022cris}
\bibfield{author}{\bibinfo{person}{Zhaoqing Wang}, \bibinfo{person}{Yu Lu}, \bibinfo{person}{Qiang Li}, \bibinfo{person}{Xunqiang Tao}, \bibinfo{person}{Yandong Guo}, \bibinfo{person}{Mingming Gong}, {and} \bibinfo{person}{Tongliang Liu}.} \bibinfo{year}{2022}\natexlab{}.
\newblock \showarticletitle{Cris: Clip-driven referring image segmentation}. In \bibinfo{booktitle}{\emph{Proceedings of the IEEE/CVF conference on computer vision and pattern recognition}}. \bibinfo{pages}{11686--11695}.
\newblock


\bibitem[Wei et~al\mbox{.}(2024)]%
        {wei2024instructseg}
\bibfield{author}{\bibinfo{person}{Cong Wei}, \bibinfo{person}{Yujie Zhong}, \bibinfo{person}{Haoxian Tan}, \bibinfo{person}{Yingsen Zeng}, \bibinfo{person}{Yong Liu}, \bibinfo{person}{Zheng Zhao}, {and} \bibinfo{person}{Yujiu Yang}.} \bibinfo{year}{2024}\natexlab{}.
\newblock \showarticletitle{InstructSeg: Unifying Instructed Visual Segmentation with Multi-modal Large Language Models}.
\newblock \bibinfo{journal}{\emph{arXiv preprint arXiv:2412.14006}} (\bibinfo{year}{2024}).
\newblock


\bibitem[Wu et~al\mbox{.}(2024)]%
        {wu2024visionllm}
\bibfield{author}{\bibinfo{person}{Jiannan Wu}, \bibinfo{person}{Muyan Zhong}, \bibinfo{person}{Sen Xing}, \bibinfo{person}{Zeqiang Lai}, \bibinfo{person}{Zhaoyang Liu}, \bibinfo{person}{Wenhai Wang}, \bibinfo{person}{Zhe Chen}, \bibinfo{person}{Xizhou Zhu}, \bibinfo{person}{Lewei Lu}, \bibinfo{person}{Tong Lu}, {et~al\mbox{.}}} \bibinfo{year}{2024}\natexlab{}.
\newblock \showarticletitle{VisionLLM v2: An End-to-End Generalist Multimodal Large Language Model for Hundreds of Vision-Language Tasks}.
\newblock \bibinfo{journal}{\emph{arXiv preprint arXiv:2406.08394}} (\bibinfo{year}{2024}).
\newblock


\bibitem[Xia et~al\mbox{.}(2024)]%
        {xia2024gsva}
\bibfield{author}{\bibinfo{person}{Zhuofan Xia}, \bibinfo{person}{Dongchen Han}, \bibinfo{person}{Yizeng Han}, \bibinfo{person}{Xuran Pan}, \bibinfo{person}{Shiji Song}, {and} \bibinfo{person}{Gao Huang}.} \bibinfo{year}{2024}\natexlab{}.
\newblock \showarticletitle{Gsva: Generalized segmentation via multimodal large language models}. In \bibinfo{booktitle}{\emph{Proceedings of the IEEE/CVF Conference on Computer Vision and Pattern Recognition}}. \bibinfo{pages}{3858--3869}.
\newblock


\bibitem[Yan et~al\mbox{.}(2024)]%
        {yan2024visa}
\bibfield{author}{\bibinfo{person}{Cilin Yan}, \bibinfo{person}{Haochen Wang}, \bibinfo{person}{Shilin Yan}, \bibinfo{person}{Xiaolong Jiang}, \bibinfo{person}{Yao Hu}, \bibinfo{person}{Guoliang Kang}, \bibinfo{person}{Weidi Xie}, {and} \bibinfo{person}{Efstratios Gavves}.} \bibinfo{year}{2024}\natexlab{}.
\newblock \showarticletitle{Visa: Reasoning video object segmentation via large language models}. In \bibinfo{booktitle}{\emph{European Conference on Computer Vision}}. Springer, \bibinfo{pages}{98--115}.
\newblock


\bibitem[Yang et~al\mbox{.}(2024b)]%
        {yang2024qwen2}
\bibfield{author}{\bibinfo{person}{An Yang}, \bibinfo{person}{Baosong Yang}, \bibinfo{person}{Binyuan Hui}, \bibinfo{person}{Bo Zheng}, \bibinfo{person}{Bowen Yu}, \bibinfo{person}{Chang Zhou}, \bibinfo{person}{Chengpeng Li}, \bibinfo{person}{Chengyuan Li}, \bibinfo{person}{Dayiheng Liu}, \bibinfo{person}{Fei Huang}, {et~al\mbox{.}}} \bibinfo{year}{2024}\natexlab{b}.
\newblock \showarticletitle{Qwen2 technical report}.
\newblock \bibinfo{journal}{\emph{arXiv preprint arXiv:2407.10671}} (\bibinfo{year}{2024}).
\newblock


\bibitem[Yang et~al\mbox{.}(2023)]%
        {yang2023lisa++}
\bibfield{author}{\bibinfo{person}{Senqiao Yang}, \bibinfo{person}{Tianyuan Qu}, \bibinfo{person}{Xin Lai}, \bibinfo{person}{Zhuotao Tian}, \bibinfo{person}{Bohao Peng}, \bibinfo{person}{Shu Liu}, {and} \bibinfo{person}{Jiaya Jia}.} \bibinfo{year}{2023}\natexlab{}.
\newblock \showarticletitle{LISA++: An Improved Baseline for Reasoning Segmentation with Large Language Model}.
\newblock \bibinfo{journal}{\emph{arXiv preprint arXiv:2312.17240}} (\bibinfo{year}{2023}).
\newblock


\bibitem[Yang et~al\mbox{.}(2024a)]%
        {yang2024empowering}
\bibfield{author}{\bibinfo{person}{Yuqi Yang}, \bibinfo{person}{Peng-Tao Jiang}, \bibinfo{person}{Jing Wang}, \bibinfo{person}{Hao Zhang}, \bibinfo{person}{Kai Zhao}, \bibinfo{person}{Jinwei Chen}, {and} \bibinfo{person}{Bo Li}.} \bibinfo{year}{2024}\natexlab{a}.
\newblock \showarticletitle{Empowering Segmentation Ability to Multi-modal Large Language Models}.
\newblock \bibinfo{journal}{\emph{arXiv preprint arXiv:2403.14141}} (\bibinfo{year}{2024}).
\newblock


\bibitem[Yang et~al\mbox{.}(2022)]%
        {yang2022lavt}
\bibfield{author}{\bibinfo{person}{Zhao Yang}, \bibinfo{person}{Jiaqi Wang}, \bibinfo{person}{Yansong Tang}, \bibinfo{person}{Kai Chen}, \bibinfo{person}{Hengshuang Zhao}, {and} \bibinfo{person}{Philip~HS Torr}.} \bibinfo{year}{2022}\natexlab{}.
\newblock \showarticletitle{Lavt: Language-aware vision transformer for referring image segmentation}. In \bibinfo{booktitle}{\emph{Proceedings of the IEEE/CVF Conference on Computer Vision and Pattern Recognition}}. \bibinfo{pages}{18155--18165}.
\newblock


\bibitem[Yao et~al\mbox{.}(2023)]%
        {yao2023dual}
\bibfield{author}{\bibinfo{person}{Ting Yao}, \bibinfo{person}{Yehao Li}, \bibinfo{person}{Yingwei Pan}, \bibinfo{person}{Yu Wang}, \bibinfo{person}{Xiao-Ping Zhang}, {and} \bibinfo{person}{Tao Mei}.} \bibinfo{year}{2023}\natexlab{}.
\newblock \showarticletitle{Dual vision transformer}.
\newblock \bibinfo{journal}{\emph{IEEE transactions on pattern analysis and machine intelligence}} \bibinfo{volume}{45}, \bibinfo{number}{9} (\bibinfo{year}{2023}), \bibinfo{pages}{10870--10882}.
\newblock


\bibitem[Ye et~al\mbox{.}(2023)]%
        {ye2023ureader}
\bibfield{author}{\bibinfo{person}{Jiabo Ye}, \bibinfo{person}{Anwen Hu}, \bibinfo{person}{Haiyang Xu}, \bibinfo{person}{Qinghao Ye}, \bibinfo{person}{Ming Yan}, \bibinfo{person}{Guohai Xu}, \bibinfo{person}{Chenliang Li}, \bibinfo{person}{Junfeng Tian}, \bibinfo{person}{Qi Qian}, \bibinfo{person}{Ji Zhang}, {et~al\mbox{.}}} \bibinfo{year}{2023}\natexlab{}.
\newblock \showarticletitle{Ureader: Universal ocr-free visually-situated language understanding with multimodal large language model}.
\newblock \bibinfo{journal}{\emph{arXiv preprint arXiv:2310.05126}} (\bibinfo{year}{2023}).
\newblock


\bibitem[Zhang et~al\mbox{.}(2023)]%
        {zhang2023next}
\bibfield{author}{\bibinfo{person}{Ao Zhang}, \bibinfo{person}{Yuan Yao}, \bibinfo{person}{Wei Ji}, \bibinfo{person}{Zhiyuan Liu}, {and} \bibinfo{person}{Tat-Seng Chua}.} \bibinfo{year}{2023}\natexlab{}.
\newblock \showarticletitle{Next-chat: An lmm for chat, detection and segmentation}.
\newblock \bibinfo{journal}{\emph{arXiv preprint arXiv:2311.04498}} (\bibinfo{year}{2023}).
\newblock


\bibitem[Zhang et~al\mbox{.}(2022)]%
        {zhang2022dino}
\bibfield{author}{\bibinfo{person}{Hao Zhang}, \bibinfo{person}{Feng Li}, \bibinfo{person}{Shilong Liu}, \bibinfo{person}{Lei Zhang}, \bibinfo{person}{Hang Su}, \bibinfo{person}{Jun Zhu}, \bibinfo{person}{Lionel~M Ni}, {and} \bibinfo{person}{Heung-Yeung Shum}.} \bibinfo{year}{2022}\natexlab{}.
\newblock \showarticletitle{Dino: Detr with improved denoising anchor boxes for end-to-end object detection}.
\newblock \bibinfo{journal}{\emph{arXiv preprint arXiv:2203.03605}} (\bibinfo{year}{2022}).
\newblock


\bibitem[Zhang et~al\mbox{.}(2024)]%
        {zhang2024omg}
\bibfield{author}{\bibinfo{person}{Tao Zhang}, \bibinfo{person}{Xiangtai Li}, \bibinfo{person}{Hao Fei}, \bibinfo{person}{Haobo Yuan}, \bibinfo{person}{Shengqiong Wu}, \bibinfo{person}{Shunping Ji}, \bibinfo{person}{Chen~Change Loy}, {and} \bibinfo{person}{Shuicheng Yan}.} \bibinfo{year}{2024}\natexlab{}.
\newblock \showarticletitle{Omg-llava: Bridging image-level, object-level, pixel-level reasoning and understanding}.
\newblock \bibinfo{journal}{\emph{arXiv preprint arXiv:2406.19389}} (\bibinfo{year}{2024}).
\newblock


\bibitem[Zheng et~al\mbox{.}(2023)]%
        {zheng2023judging}
\bibfield{author}{\bibinfo{person}{Lianmin Zheng}, \bibinfo{person}{Wei-Lin Chiang}, \bibinfo{person}{Ying Sheng}, \bibinfo{person}{Siyuan Zhuang}, \bibinfo{person}{Zhanghao Wu}, \bibinfo{person}{Yonghao Zhuang}, \bibinfo{person}{Zi Lin}, \bibinfo{person}{Zhuohan Li}, \bibinfo{person}{Dacheng Li}, \bibinfo{person}{Eric Xing}, {et~al\mbox{.}}} \bibinfo{year}{2023}\natexlab{}.
\newblock \showarticletitle{Judging llm-as-a-judge with mt-bench and chatbot arena}.
\newblock \bibinfo{journal}{\emph{Advances in Neural Information Processing Systems}}  \bibinfo{volume}{36} (\bibinfo{year}{2023}), \bibinfo{pages}{46595--46623}.
\newblock


\bibitem[Zheng et~al\mbox{.}(2024)]%
        {zheng2024villa}
\bibfield{author}{\bibinfo{person}{Rongkun Zheng}, \bibinfo{person}{Lu Qi}, \bibinfo{person}{Xi Chen}, \bibinfo{person}{Yi Wang}, \bibinfo{person}{Kun Wang}, \bibinfo{person}{Yu Qiao}, {and} \bibinfo{person}{Hengshuang Zhao}.} \bibinfo{year}{2024}\natexlab{}.
\newblock \showarticletitle{ViLLa: Video Reasoning Segmentation with Large Language Model}.
\newblock \bibinfo{journal}{\emph{arXiv preprint arXiv:2407.14500}} (\bibinfo{year}{2024}).
\newblock


\bibitem[Zhou et~al\mbox{.}(2017)]%
        {zhou2017scene}
\bibfield{author}{\bibinfo{person}{Bolei Zhou}, \bibinfo{person}{Hang Zhao}, \bibinfo{person}{Xavier Puig}, \bibinfo{person}{Sanja Fidler}, \bibinfo{person}{Adela Barriuso}, {and} \bibinfo{person}{Antonio Torralba}.} \bibinfo{year}{2017}\natexlab{}.
\newblock \showarticletitle{Scene parsing through ade20k dataset}. In \bibinfo{booktitle}{\emph{Proceedings of the IEEE conference on computer vision and pattern recognition}}. \bibinfo{pages}{633--641}.
\newblock


\bibitem[Zhu et~al\mbox{.}(2023b)]%
        {zhu2023egoobjects}
\bibfield{author}{\bibinfo{person}{Chenchen Zhu}, \bibinfo{person}{Fanyi Xiao}, \bibinfo{person}{Andr{\'e}s Alvarado}, \bibinfo{person}{Yasmine Babaei}, \bibinfo{person}{Jiabo Hu}, \bibinfo{person}{Hichem El-Mohri}, \bibinfo{person}{Sean Culatana}, \bibinfo{person}{Roshan Sumbaly}, {and} \bibinfo{person}{Zhicheng Yan}.} \bibinfo{year}{2023}\natexlab{b}.
\newblock \showarticletitle{Egoobjects: A large-scale egocentric dataset for fine-grained object understanding}. In \bibinfo{booktitle}{\emph{Proceedings of the IEEE/CVF International Conference on Computer Vision}}. \bibinfo{pages}{20110--20120}.
\newblock


\bibitem[Zhu et~al\mbox{.}(2023a)]%
        {zhu2023minigpt}
\bibfield{author}{\bibinfo{person}{Deyao Zhu}, \bibinfo{person}{Jun Chen}, \bibinfo{person}{Xiaoqian Shen}, \bibinfo{person}{Xiang Li}, {and} \bibinfo{person}{Mohamed Elhoseiny}.} \bibinfo{year}{2023}\natexlab{a}.
\newblock \showarticletitle{Minigpt-4: Enhancing vision-language understanding with advanced large language models}.
\newblock \bibinfo{journal}{\emph{arXiv preprint arXiv:2304.10592}} (\bibinfo{year}{2023}).
\newblock


\bibitem[Zou et~al\mbox{.}(2023)]%
        {zou2023generalized}
\bibfield{author}{\bibinfo{person}{Xueyan Zou}, \bibinfo{person}{Zi-Yi Dou}, \bibinfo{person}{Jianwei Yang}, \bibinfo{person}{Zhe Gan}, \bibinfo{person}{Linjie Li}, \bibinfo{person}{Chunyuan Li}, \bibinfo{person}{Xiyang Dai}, \bibinfo{person}{Harkirat Behl}, \bibinfo{person}{Jianfeng Wang}, \bibinfo{person}{Lu Yuan}, {et~al\mbox{.}}} \bibinfo{year}{2023}\natexlab{}.
\newblock \showarticletitle{Generalized decoding for pixel, image, and language}. In \bibinfo{booktitle}{\emph{Proceedings of the IEEE/CVF Conference on Computer Vision and Pattern Recognition}}. \bibinfo{pages}{15116--15127}.
\newblock


\bibitem[Zou et~al\mbox{.}(2024)]%
        {zou2024segment}
\bibfield{author}{\bibinfo{person}{Xueyan Zou}, \bibinfo{person}{Jianwei Yang}, \bibinfo{person}{Hao Zhang}, \bibinfo{person}{Feng Li}, \bibinfo{person}{Linjie Li}, \bibinfo{person}{Jianfeng Wang}, \bibinfo{person}{Lijuan Wang}, \bibinfo{person}{Jianfeng Gao}, {and} \bibinfo{person}{Yong~Jae Lee}.} \bibinfo{year}{2024}\natexlab{}.
\newblock \showarticletitle{Segment everything everywhere all at once}.
\newblock \bibinfo{journal}{\emph{Advances in Neural Information Processing Systems}}  \bibinfo{volume}{36} (\bibinfo{year}{2024}).
\newblock


\end{thebibliography}

\end{document}